\documentclass{article}

\usepackage[main, final, nonatbib]{neurips_2026}

\usepackage[utf8]{inputenc} %
\usepackage[T1]{fontenc}    %
\usepackage{hyperref}       %
\usepackage{url}            %
\usepackage{booktabs}       %
\usepackage{amsfonts}       %
\usepackage{nicefrac}       %
\usepackage{microtype}      %
\usepackage{xcolor}         %

\usepackage{enumitem}
\usepackage{amsmath}
\usepackage[table]{xcolor}  %
\usepackage{adjustbox}
\usepackage{multirow}
\usepackage{comment}
\usepackage{wrapfig}
\usepackage{algorithm}
\usepackage{algpseudocode}

\title{HeMeR: Heterogeneous Memory Reconciliation for Embodied Agents via Structured KV Reuse}

\usepackage[style=numeric, sorting=none]{biblatex}
\newcommand{\ourmodel}{\textsc{HeMeR}}

\author{%
Saehun Chun$^{1}$, Wonje Choi$^{1}$, Jinwoo Jang$^{1}$, Taeyoon Kwack$^{2}$, Honguk Woo$^{1,2}$\thanks{Honguk Woo is the corresponding author.} \\
$^{1}$Department of Computer Science and Engineering, Sungkyunkwan University \\
$^{2}$Department of Artificial Intelligence, Sungkyunkwan University \\
\texttt{\{thousae,wjchoi1995,jinustar,njj05043,hwoo\}@skku.edu} \\
}

\begin{document}
\raggedbottom
\maketitle

\begin{abstract}
Embodied agents make decisions by combining information from heterogeneous memory modules, including declarative memory for explicit knowledge, procedural memory for learned routines, and working memory for current observations.
During interaction, the memory contents required for decision making change with the agent's state, goal, and task progress, while many contents remain shared across consecutive decisions.
However, existing memory modules are typically accessed through separate interfaces, so each updated memory context is often processed as a new input to the foundation model.
This prevents selective key-value (KV) cache reuse for unchanged memory contents and forces full KV cache recomputation even when much of the memory context remains valid.
We introduce $\ourmodel$, an inference framework that performs KV reconciliation between dynamically updated memory contexts and reusable KV states.
$\ourmodel$ tracks declarative, procedural, and working memory contents across action decisions, preserves KV states for unchanged contents, and reconciles only the KV states affected by memory-content updates.
This enables efficient and stable inference as memory contents evolve during interaction.
We evaluate $\ourmodel$ on ALFWorld, Habitat, and real-world robotic deployments requiring coordinated use of heterogeneous memory contents.
In the real-world setting, $\ourmodel$ achieves $83.20\%$ task success and reduces per-step inference time from $2.94$ to $1.87$ seconds compared with MemRL, while preserving compatible KV states as task progress updates the memory context.
\end{abstract}

\section{Introduction}\label{sec:intro}
Embodied agents increasingly build on foundation models, using language or vision-language models to interpret observations, reason about goals, and select actions~\cite{embodied:saycan,base:react,embodied:llmplanner,embodied:inner,embodied:progprompt,embodied:voyager}.
However, embodied tasks often require information beyond immediate observations, including explicit environmental knowledge, learned behavioral routines, and ongoing task progress~\cite{embodied:inner,embodied:progprompt,embodied:voyager,embodied:reflexion}.
Accordingly, recent studies augment foundation models with memory modules that store, retrieve, and update task-relevant memory contents across interactions~\cite{embodied:voyager,embodied:reflexion,method:genagent,serve:memgpt,mem:memorybank}.
These modules help agents preserve continuity over long-horizon tasks, but also make each action decision depend on memory contents that change with accumulated experience, current observations, and task progress.

Recent memory-augmented agents use heterogeneous memory modules that maintain distinct forms of memory content for embodied decision making~\cite{base:react,embodied:voyager,method:genagent,embodied:coala}.
A declarative memory module maintains explicit content about environmental objects, relations, and task conditions, often as structured representations such as symbolic facts~\cite{method:genagent,serve:memgpt,mem:amem,mem:nesyc}.
A procedural memory module maintains learned behavioral routines that guide future actions~\cite{embodied:progprompt,embodied:voyager,mem:procedural,mem:nesypr}, including parametric representations such as neural adapters~\cite{adapt:lora}.
A working memory module maintains content about the current task state, including recent observations, intermediate decisions, and goal progress~\cite{base:react,embodied:inner,mem:working}.
Although these modules are often studied through separate storage, retrieval, and update mechanisms, action selection requires their contents to be composed into a memory context.
The resulting memory context is processed by the foundation model to form transformer KV states during inference~\cite{mem:kv,serve:vllm}.
As the agent receives new observations and progresses through the task, the memory context must be updated while many memory contents remain unchanged across consecutive decisions.

\begin{figure*}[t]
\centering
\includegraphics[width=0.99\linewidth]{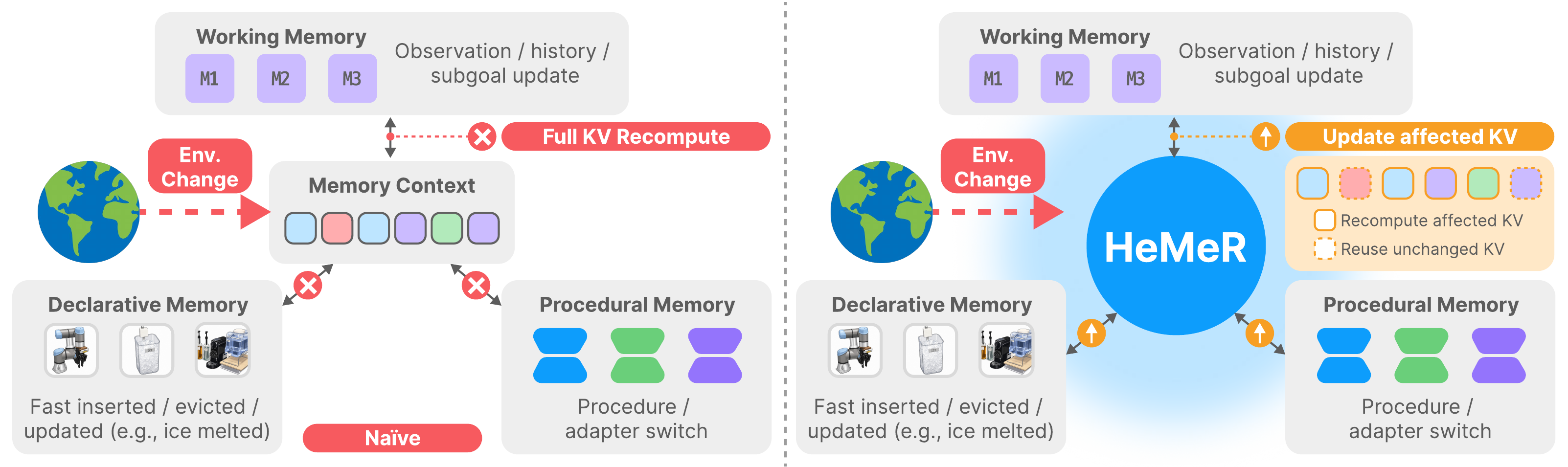}
\caption{
\textbf{Motivation of $\ourmodel$.}
Environment changes can trigger memory-content updates, including procedural changes that switch the active adapter.
A naive implementation recomputes the full KV cache, whereas $\ourmodel$ updates affected KV ranges and reuses compatible KV states.
}
\label{fig:motivation}
\end{figure*}

This modular separation is well motivated, since each memory module serves a distinct role in embodied decision making~\cite{embodied:coala, mem:procedural, mem:working, mem:episodic}.
However, after memory contents from multiple memory modules are composed into a single memory context, the foundation model must process this memory context to form the hidden representations and KV states used for action selection~\cite{mem:kv,serve:vllm}.
As illustrated in Figure~\ref{fig:motivation}, changes in the environment can trigger updates to declarative and working memory contents and procedural adapter switches, thereby changing the memory context across decision steps.
Such changes can make some previously cached KV states incompatible with the updated memory context, while others may remain reusable~\cite{serve:cacheblend, serve:kvlink, serve:cacheslide, pos:rope, silo:lragent, serve:alora}.
Existing memory-augmented agents mainly focus on what to store, retrieve, and update, but do not explicitly address how to reconcile updated memory contexts with reusable KV states during foundation model inference~\cite{serve:memgpt, mem:memorybank, mem:amem, serve:memos, mem:titans}.
Without such reconciliation, a naive implementation processes each updated memory context as a new input and recomputes the full KV cache, causing redundant computation during long-horizon embodied interaction~\cite{serve:kvlink,serve:epic,serve:cachegen,serve:lmcache}.

To address this problem, we introduce $\ourmodel$, a memory-serving framework for embodied agents that reconciles dynamically updated memory contexts with reusable KV states during foundation model inference.
Although declarative, procedural, and working memory serve different roles, their effects on foundation model inference are ultimately reflected in the KV states used for action selection.
Rather than treating each memory context as a flat token sequence, $\ourmodel$ represents it as a sequence of structured memory chunks associated with corresponding KV ranges, providing a common basis for reconciling heterogeneous memory updates with cached KV states.
$\ourmodel$ realizes this reconciliation through two complementary mechanisms, \textit{Chunk-Level KV Reconciliation} for content-level changes and rearrangements within the memory context, and \textit{Adapter-Stable KV Reconciliation} for procedural changes that alter the active adapter.

Specifically, \textit{Chunk-Level KV Reconciliation} handles updates to declarative and working memory contents by mapping content-level changes and rearrangements to their corresponding KV ranges.
$\ourmodel$ identifies reusable KV ranges by tracking changes in memory contents and their dependencies, selectively recomputing only the affected ranges.
When an unchanged memory chunk shifts position, RoPE-based remapping preserves positional consistency, allowing its KV states to be reused~\cite{pos:rope}.
\textit{Adapter-Stable KV Reconciliation} handles procedural memory updates that alter the active adapter.
Because KV states computed under a previous adapter may become incompatible even when other memory contents remain unchanged, $\ourmodel$ treats the adapter configuration as part of the KV compatibility condition, selectively recomputing affected KV ranges while preserving compatible KV states across adapter switches.
By jointly reconciling memory content updates and procedural adapter changes without rebuilding the full KV cache, $\ourmodel$ supports efficient and stable inference over long-horizon embodied tasks.

We evaluate $\ourmodel$ on unseen simulated tasks from ALFWorld~\cite{ben:alfworld} and EmbodiedBench Habitat~\cite{ben:embodiedbench,ben:habitat}, where evaluation episodes require the agent to combine memory contents across different situations rather than rely on a directly applicable memory content.
We further evaluate $\ourmodel$ on real-world embodied tasks in a UR7e manipulation setup, requiring the agent to coordinate heterogeneous memory contents as procedures and task states change.
We highlight task success rate and inference time per decision step, with each step corresponding to one action selection.
On ALFWorld, $\ourmodel$ achieves a $71.43\%$ task success rate with $0.18$ seconds per decision step, compared with a $52.86\%$ task success rate and $0.90$ seconds per decision step for MemRL~\cite{mem:memrl}, the strongest baseline in task success (Table~\ref{tab:main:sim}).
On EmbodiedBench Habitat, $\ourmodel$ achieves a $76.27\%$ task success rate with $1.82$ seconds per decision step, compared with a $61.80\%$ task success rate and $2.68$ seconds per decision step for MemRL (Table~\ref{tab:main:sim}).
In the real-world setting, $\ourmodel$ achieves an 83.20\% goal-conditioned success rate with 1.87 seconds per decision step, compared with a 64.28\% goal-conditioned success rate and 2.94 seconds per decision step for MemRL (Table~\ref{tab:main:real-world}).
These results show that KV reconciliation improves task robustness and inference efficiency as memory contexts change across embodied decision steps.

We make the following contributions.
\begin{itemize}[leftmargin=*, itemsep=0pt, topsep=0pt]
    \item We formulate heterogeneous memory use as a KV reconciliation problem for foundation model-based embodied agents, where memory contents from declarative, procedural, and working memory modules change across action decisions while many contents remain reusable.
    \item We introduce $\ourmodel$, a heterogeneous memory reconciliation framework that anchors memory contents as typed chunks with KV ranges and selectively reuses compatible KV states.
    \item We develop chunk-level and adapter-stable KV reconciliation mechanisms that update only affected KV ranges, preserve positional consistency under memory-content updates, and maintain compatible KV states across procedural adapter switches.
    \item We evaluate $\ourmodel$ on ALFWorld, Habitat, and real-world robotic deployments, showing that it improves task success and reduces per-step inference time over strong baselines.
\end{itemize}

\section{Related work}\label{sec:related}

\paragraph{Foundation Model-based Embodied Control.}
Recent studies use language and vision-language foundation models as general-purpose decision modules for embodied agents, enabling them to interpret observations, reason about goals, and select actions in interactive environments~\cite{embodied:saycan, embodied:llmplanner, embodied:inner, embodied:progprompt, embodied:voyager, embodied:ecot}.
These agents are further augmented with memory modules to store and retrieve task-relevant memory contents across decision steps and adapt action selection over long-horizon interactions~\cite{embodied:reflexion, mem:memorybank, embodied:expel, mem:memrl, embodied:jarvis1}.
$\ourmodel$ follows this line of work, but focuses on the inference problem that arises when memory contents from multiple memory modules are composed into a dynamically updated memory context that must be reconciled with reusable KV states across action decisions.

\paragraph{Memory-Augmented Embodied Agents.}
Recent memory-augmented embodied agents introduce specialized memory modules for different roles in decision making.
Declarative memory modules maintain explicit content about environmental objects, relations, and task conditions~\cite{method:genagent, mem:memorybank, mem:amem, mem:nesyc}.
Procedural memory modules maintain learned behavioral routines, including reusable procedures and skills~\cite{embodied:progprompt, embodied:voyager, mem:procedural, mem:nesypr, mem:memp, mem:automanual, embodied:rpms, silo:lragent}.
Working memory modules maintain memory content about recent observations, intermediate decisions, and goal progress during the current interaction~\cite{base:react, embodied:inner, embodied:coala, mem:working}.
These memory modules help embodied agents preserve task-relevant memory content beyond immediate observations and adapt action selection over time.
However, existing studies mainly focus on how each memory module stores, retrieves, or updates its own memory content for a role-specific objective.
In contrast, $\ourmodel$ jointly handles memory contents from declarative, procedural, and working memory modules once they are composed into a single memory context, reconciling their updates with reusable KV states across action decisions.

\paragraph{Efficient KV Reuse.}
KV caching is a standard mechanism for efficient autoregressive inference, in which previously computed KV states are stored and reused across decoding steps~\cite{mem:kv, serve:vllm, serve:orca}.
Prior work on KV reuse reduces inference cost by reusing cached KV states for repeated prefixes, shared prompts, or recurring context segments~\cite{serve:epic,kv:promptcache,mem:fcgraft}.
Recent systems further exploit structured generation programs, retrieved context chunks, position-independent context segments, or embodied reasoning traces to reuse computation across related inference requests or decision steps~\cite{serve:cacheblend, serve:kvlink, serve:cacheslide, serve:sglang, embodied:fastecot, embodied:keep}.
Complementary work reduces KV cache storage requirements, transfer costs, or loading overhead through eviction, compression, streaming, or cache-layer management~\cite{serve:cachegen, serve:lmcache, serve:h2o, serve:streamingllm}.
In parallel, multi-adapter serving systems improve the efficiency of models augmented with LoRA adapters by batching requests, switching adapters, or reusing KV states across adapters during inference~\cite{adapt:lora, silo:lragent, serve:alora, serve:slora, serve:punica, serve:dlora, adapt:libra}.
Building on these advances, $\ourmodel$ integrates KV reuse and multi-adapter serving techniques within a memory-serving framework to efficiently reuse KV states as memory contexts change across action decisions.

\section{Problem formulation}\label{sec:problem}
We consider a foundation model-based embodied agent that acts over a sequence of decision steps.
At step $t$, the agent receives an observation $o_t$, maintains a task goal $g$, and selects an action $a_t$ using a foundation model $F_\theta$.
The agent is augmented with heterogeneous memory modules
\begin{equation}
    \mathcal{M} = \{\mathcal{M}^{\mathrm{dec}}, \mathcal{M}^{\mathrm{proc}}, \mathcal{M}^{\mathrm{work}}\},
\end{equation}
corresponding to declarative, procedural, and working memory modules.
A memory module stores and retrieves information, and each item it returns is a memory content.
For each module $\mathcal{M}^m$, let $R^m$ denote its retrieval function and $\mathcal{R}_t^m$ the memory contents obtained at step $t$:
\begin{equation}
    \mathcal{R}_t^m = R^m(o_t, g, h_t), \quad m \in \{\mathrm{dec}, \mathrm{proc}, \mathrm{work}\}.
\end{equation}
Here, $h_t$ denotes the interaction history available at step $t$.

For readability, we denote the declarative, procedural, and working-memory contents as
\begin{equation}
    D_t = \mathcal{R}_t^{\mathrm{dec}},
    \quad
    P_t = \mathcal{R}_t^{\mathrm{proc}},
    \quad
    W_t = \mathcal{R}_t^{\mathrm{work}}.
\end{equation}
A memory context is formed by composing multiple memory contents to support action selection.
The function $\Gamma$ defines the step-specific memory context $X_t$, which the foundation model uses together with the observation and task goal to generate an action $a_t$:
\begin{equation}
    X_t = \Gamma(D_t, P_t, W_t), \qquad a_t \sim F_\theta(\cdot \mid X_t, o_t, g).
\end{equation}
Let $\mathcal{D}$ denote the distribution of tasks encountered by the agent.
Across decision steps, $X_t$ may change as memory contents are inserted, evicted, reordered, or associated with different procedural adapters.
Let $\Pi$ denote a serving policy that reconciles the previous KV cache with the current memory context $X_t$.
$\Pi$ determines which KV states are reused or recomputed during action generation.
Our objective is to maximize task success while minimizing inference cost:
\begin{equation}
    \Pi^\star =
    \underset{\Pi}{\operatorname{arg\max}}
    \underset{\tau \sim \mathcal{D}}{\mathbb E}
    \left[
        \mathrm{Succ}(\tau;\Pi) - \lambda \sum_{t=1}^{T} \mathrm{Cost}\!\left(\Pi, X_{t-1}, X_t\right)
    \right].
\end{equation}
Here, $\tau$ denotes a task instance sampled from $\mathcal{D}$,
$\mathrm{Succ}(\tau;\Pi)$ measures whether $\tau$ is successfully completed under $\Pi$,
$\mathrm{Cost}(\Pi, X_{t-1}, X_t)$ measures the inference computation incurred under $\Pi$ when processing the memory context update from $X_{t-1}$ to $X_t$, and $\lambda$ controls the trade-off between task success and inference cost.

\section{Method}\label{sec:method}
\begin{figure*}[!t]
\centering
\includegraphics[width=0.99\linewidth]{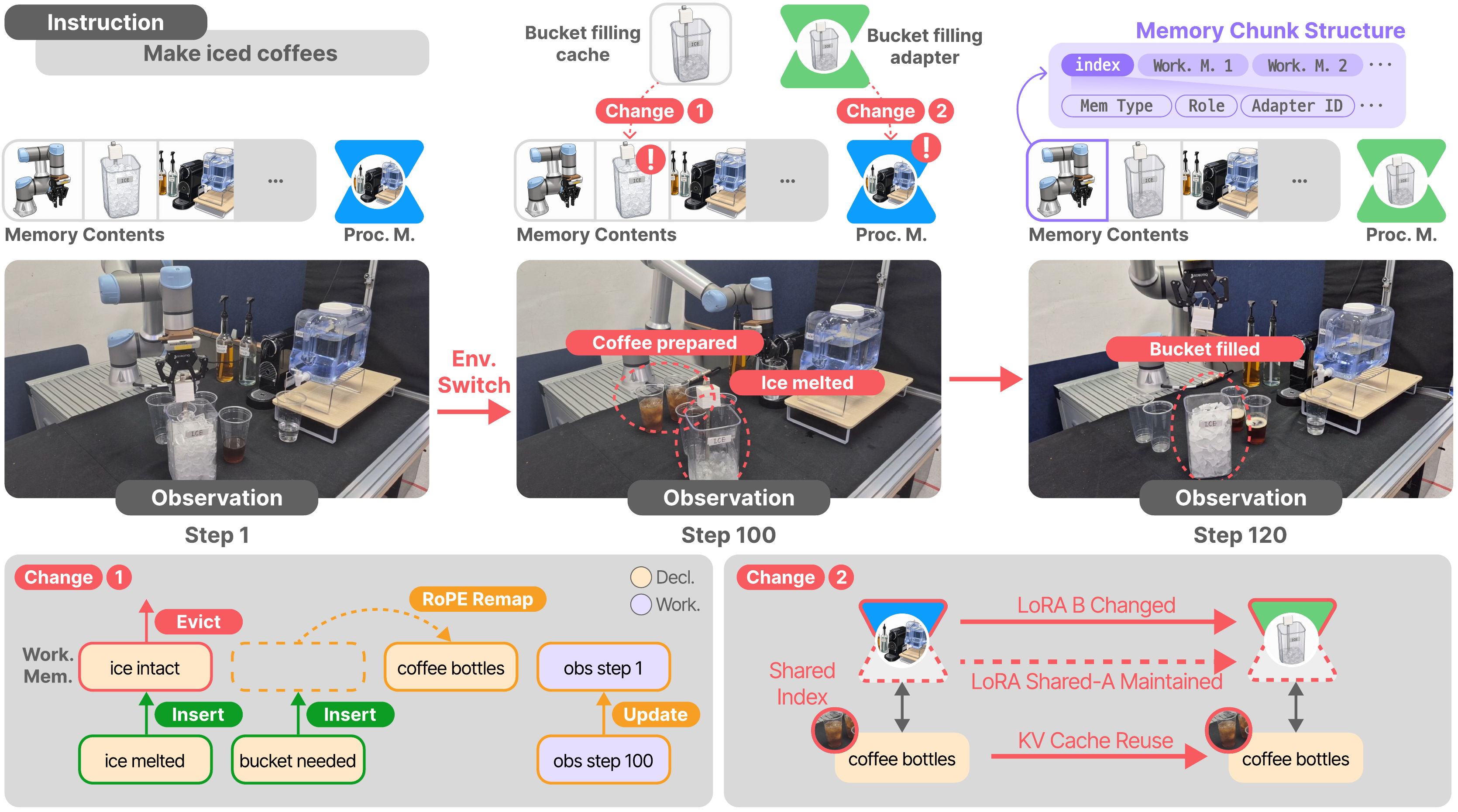}
\caption{
$\ourmodel$ overview.
During long-horizon inference, new observations can modify the memory context and switch the active procedural adapter.
$\ourmodel$ indexes memory chunks, updates affected KV ranges, and preserves compatible cache states across decision steps.
}
\label{fig:fig2}
\end{figure*}

\subsection{Overall Architecture}\label{sec:method:overall}

We present $\ourmodel$, a heterogeneous memory reconciliation framework that enables foundation model-based embodied agents to process changing memory contexts without rebuilding the full KV cache, as illustrated in Figure~\ref{fig:fig2}.
At each decision step, the agent obtains memory contents from multiple memory modules and composes them into a single memory context, which the foundation model uses together with the current observation and task goal to generate the next action.
During interaction, new observations may introduce memory contents, outdated contents may be removed, existing contents may be reordered, and the active procedural adapter may switch.
To handle these changes, $\ourmodel$ uses a memory schema to anchor each memory content as a typed memory chunk with an associated KV range.
It then reconciles memory-content updates through localized cache operations while preserving compatible KV states across procedural adapter switches.

We distinguish three levels of memory representation: memory modules, memory contents, and memory contexts.
A memory module stores and retrieves a specific type of information, and each item retrieved at a decision step is a memory content.
A memory context is formed by composing multiple memory contents to support action selection.
In our setting, the agent uses declarative, procedural, and working memory modules.
Declarative memory stores explicit task and environment knowledge, such as object locations, receptacle relations, and task conditions.
Procedural memory stores reusable behavioral procedures, instantiated as LoRA adapters in our LRAgent-based implementation~\cite{silo:lragent}.
Working memory stores the current interaction state, including recent observations, intermediate decisions, subgoals, and action history.

Rather than treating the composed memory context $X_t$ defined in Sec.~\ref{sec:problem} solely as a flat token sequence, $\ourmodel$ organizes it as a sequence of memory chunks.
Each memory chunk represents a single memory content, with schema-defined metadata linking it to a contiguous KV range after tokenization and encoding.
The memory schema records each chunk's persistent ID, memory module type, semantic role, normalized content, dependency annotation, position, token span, and KV range, as well as its adapter identifier when applicable.
This index maps memory-content updates to their corresponding KV ranges, enabling $\ourmodel$ to reconcile the KV cache through localized operations on affected ranges.
The initial memory context is prefilled as a single causal sequence.
At subsequent decision steps, $\ourmodel$ prefills newly inserted memory chunks and selectively recomputes affected KV ranges at their positions in the current memory context, recording the resulting KV ranges in the memory schema.

\subsection{Chunk-Level KV Reconciliation}
\label{sec:method:kv}

Chunk-level KV reconciliation aligns the previous KV cache with the memory contents in $X_t$.
As the agent progresses, memory contents may be inserted, evicted, or reordered, potentially altering token positions and dependencies in $X_t$.
Naive KV reuse is therefore unreliable, while rebuilding the full KV cache can introduce redundant computation when many memory contents remain unchanged.

\paragraph{Schema-Guided Reconciliation.}
To address this trade-off, $\ourmodel$ uses the chunk-level index to compare memory chunks across consecutive decision steps.
Chunks in $X_{t-1}$ and $X_t$ are first matched by persistent chunk IDs, regardless of their positions or token spans.
Matched chunks are eligible for KV reuse only when their memory module type, semantic role, dependency annotation, normalized content, and adapter identifier (when applicable) agree.
Otherwise, their KV ranges are locally recomputed.
Given two consecutive memory contexts, $\ourmodel$ computes a schema-guided reconciliation plan
\begin{equation}
    \Delta_t = \mathrm{Diff}_{\mathcal{S}}(X_{t-1}, X_t),
\end{equation}
where $\mathrm{Diff}_{\mathcal{S}}$ matches chunks by persistent IDs and checks schema compatibility.
Matched chunks with changed positions are reordered, chunks only in $X_t$ are inserted, and chunks only in $X_{t-1}$ are evicted.
The index maps these operations to KV ranges, so $\ourmodel$ updates only affected ranges.

\paragraph{KV Operations and RoPE Remapping.}
Based on the reconciliation plan $\Delta_t$, $\ourmodel$ performs the corresponding KV operations.
New memory chunks are prefilled, while the KV ranges of evicted chunks are removed.
When chunks are reordered, $\ourmodel$ relocates their KV ranges according to the current $X_t$, reusing compatible KV states when possible.
These operations allow the KV cache to be updated without full recomputation.

RoPE remapping is used only for matched chunks whose update is a position shift.
If a reusable token shifts from position $p_j$ to $p'_j$, its cached key is updated as
\begin{equation}
    \tilde{k}_{j} = R_{\theta(p'_j)}^{} R_{\theta(p_j)}^{-1} k_j,
\end{equation}
where $k_j$ denotes the cached key at position $p_j$, $R_{\theta(p)}$ denotes the RoPE rotation at position $p$, and $\tilde{k}_j$ denotes the remapped key at position $p'_j$.
This corrects the cached key's positional phase under a pure position shift.
RoPE remapping updates only the cached keys.
For multimodal backbones, we restrict RoPE-based remapping to text-side memory chunks whose schema fields remain unchanged~\cite{pos:rope}.
Visual tokens from current observations are not position-remapped and are encoded normally at each decision step.
When content, dependencies, or adapters change, $\ourmodel$ recomputes the affected KV range to avoid stale reuse.

Chunk-level KV reconciliation differs from prompt-level or segment-level cache reuse.
Prefix caching reuses only stable prefixes, Prompt Cache reuses segments computed without their preceding context, and CacheBlend-style reuse still relies on reusable segment boundaries and selective recomputation around them~\cite{serve:cacheblend, serve:kvlink, kv:promptcache}.
$\ourmodel$ makes the same class of non-prefix reuse approximation, but grounds it in schema-indexed memory chunks and reuses a KV range only when the corresponding memory content remains schema-compatible.
Thus, $\ourmodel$ avoids both naive KV reuse and full KV cache recomputation by selectively reconciling KV ranges affected by memory-content updates across decision steps.

\subsection{Adapter-Stable KV Reconciliation}
\label{sec:method:adapter}

Procedural memory introduces a second source of cache incompatibility.
In our LRAgent-based implementation, procedural memory contents correspond to reusable behavioral procedures implemented as LoRA adapters.
When the procedural memory content changes across decision steps, the active adapter may also change.
Even if declarative and working-memory contents remain unchanged, KV states computed under the previous adapter may no longer be compatible with the new adapter configuration.
This issue is particularly important in embodied tasks, where agents switch among procedures such as searching, manipulating, cleaning, heating, cooling, and placing objects.

\paragraph{Adapter-Aware Reconciliation.}
$\ourmodel$ maintains adapter identifiers as part of the chunk-level index.
For each procedural memory chunk, the index records the adapter under which its KV states were computed, and the reconciliation plan marks adapter switches as procedural updates.
When the active adapter changes, $\ourmodel$ recomputes the procedural memory chunk's KV range and those of its direct and transitive dependents under the declared dependency annotations.
Other unchanged ranges are approximately reused under the shared-attention LoRA design.
This allows adapter switches to be handled within the same indexed reconciliation process without rebuilding the full KV cache.

\paragraph{Cross-Adapter KV Compatibility.}
To stabilize cross-adapter reuse, $\ourmodel$ adopts the shared-attention LoRA design used by LRAgent. 
The rank-8 adapters are applied to the query and value projections, with a low-rank down-projection shared across adapters and adapter-specific up-projections. 
The key, output, and MLP paths remain shared, and the active adapter is applied to the entire forward sequence rather than only to procedural-memory tokens. 
This design limits, but does not eliminate, cross-adapter drift. 
Therefore, $\ourmodel$ includes the adapter identifier in its KV compatibility condition and recomputes the procedural memory chunk's KV range while preserving ranges that remain compatible across the switch.

While LRAgent primarily addresses KV cache reuse across adapter switches, $\ourmodel$ incorporates adapter compatibility into the same reconciliation process used for updates to declarative and working memory contents.
Such methods reduce overhead for procedural changes, but they do not explicitly address updates to declarative and working memory contents that arise during embodied interaction.
$\ourmodel$ jointly handles both sources of change: adapter switches from procedural memory and chunk-level reconciliation of updates to declarative and working-memory contents.
Consequently, $\ourmodel$ preserves compatible KV states across heterogeneous memory-content updates while avoiding the need to rebuild the full KV cache at every decision step.

\section{Evaluation}\label{sec:eval}

\subsection{Experimental setting}\label{sec:eval:sub1}
\paragraph{Environment.}
We evaluate on ALFWorld~\cite{ben:alfworld}, EmbodiedBench Habitat~\cite{ben:embodiedbench,ben:habitat}, and Real-world, a UR7e manipulation setup.
Each evaluation episode is unseen in the sense that no single memory fragment covers the required procedure-room combination.
Solving these tasks therefore requires composing fragmented memories across different situations rather than retrieving a directly applicable past experience.
Real-world evaluation comprises two long-horizon tasks in two tabletop environments, with five independent physical repetitions per task, yielding 10 trials per method.
Initial object positions and instruction phrasings vary across repetitions.
Benchmark configurations, task details, and evaluation procedures are provided in Appendix~\ref{app:exp}.

\paragraph{Model.}
We use two backbone models matched to the observation modality of each environment. 
For ALFWorld, a text-based environment, we use Llama-3.2-3B-Instruct~\cite{llm:llama}. 
For EmbodiedBench Habitat, a vision-based 3D environment, we use Qwen3-VL-4B-Instruct~\cite{llm:qwenvl}, which natively processes RGB observations. 
Both are served with greedy decoding (temperature$\,{=}\,0$). 
For procedural memory, we use rank-8 LoRA adapters~\cite{adapt:lora} fine-tuned on task-type-specific trajectories for each backbone.
Each procedural memory content is linked to an adapter identifier used for adapter selection and KV compatibility checking, as described in Sec.~\ref{sec:method:adapter}.

\paragraph{Baselines.}
We compare $\ourmodel$ with six baselines covering agent-level reasoning, memory use, KV reuse, and procedural adapter switching.
ReAct~\cite{base:react} tests prompting without explicit external memory, and Reflexion~\cite{embodied:reflexion} tests self-reflective feedback from prior failures.
MemRL~\cite{mem:memrl} retrieves episodic memory from previous trajectories to guide action selection, but treats the retrieved context as a conventional input across decision steps.
CacheBlend~\cite{serve:cacheblend} and KEEP~\cite{embodied:keep} evaluate segment-level and invalidation-based KV reuse, respectively.
LRAgent~\cite{silo:lragent} tests LoRA-based procedural skill switching, but does not explicitly track updates in declarative or working-memory contents.
Together, these baselines separate the effects of prompting, episodic memory retrieval, cache reuse, procedural adapter switching, and the proposed heterogeneous-memory KV reconciliation.

\paragraph{Metrics.}
We evaluate task performance, action behavior, and inference efficiency with six metrics.
SR measures binary end-to-end task success in ALFWorld and Habitat~\cite{ben:alfworld,ben:habitat}, and goal-conditioned success rate in real-world tasks.
Latency measures average wall-clock time per decision step, while TTFT isolates prefill and cache-processing latency before the first generated token, a common serving-level efficiency metric~\cite{serve:vllm,serve:cacheblend}.
Reuse ratio measures the fraction of KV states preserved after memory-content updates, following prior cache reuse studies~\cite{serve:cacheblend,serve:epic}.
Active action ratio measures the fraction of generated actions belonging to predefined task-progressing action categories~\cite{ben:alfred}.
Action diversity measures the average per-episode ratio of unique actions among all generated actions~\cite{div:roboagent}.
In the tables, active ratio and diversity ratio denote active action ratio and action diversity ratio, respectively.

\subsection{Main result}\label{sec:eval:main}
\begin{table*}[!t]
\caption{
Performance on embodied tasks with dynamic memory contexts.
$\ourmodel$ improves task success while reducing inference time through higher KV reuse and effective memory utilization.}
\label{tab:main:sim}
\centering
\begin{adjustbox}{width=.99\textwidth}
\begin{tabular}{l ccccc}
    \toprule
	\textbf{ALFWorld} & \textbf{SR (\%)} & \textbf{Latency (s/step)} & \textbf{Reuse ratio (\%)} & \textbf{Active ratio (\%)} & \textbf{Diversity ratio (\%)}\\
    \cmidrule(rl){1-1} \cmidrule(rl){2-6}
    \textbf{$\ourmodel$} & $\mathbf{71.43 \pm 2.05}$ & $\mathbf{0.18 \pm 0.02}$ & $\mathbf{75.30 \pm 4.33}$ & $\mathbf{67.69 \pm 2.27}$ & $\mathbf{71.43 \pm 3.32}$ \\
    ReAct & $47.14 \pm 6.39$ & $0.88 \pm 0.04$ & -- & $52.04 \pm 2.45$ & $64.95 \pm 8.94$ \\
    Reflexion & $44.29 \pm 5.74$ & $0.98 \pm 0.05$ & -- & $45.57 \pm 4.30$ & $64.38 \pm 6.12$ \\
    MemRL & $52.86 \pm 4.83$ & $0.90 \pm 0.04$ & -- & $46.26 \pm 5.16$ & $65.05 \pm 4.48$ \\
    CacheBlend & $48.39 \pm 4.29$ & $0.48 \pm 0.04$ & $50.30 \pm 2.33$ & $46.01 \pm 3.67$ & $58.49 \pm 2.92$ \\
    KEEP & $51.81 \pm 3.92$ & $0.46 \pm 0.06$ & $48.30 \pm 2.33$ & $50.40 \pm 4.92$ & $65.01 \pm 4.95$ \\
    LRAgent & $46.43 \pm 3.19$ & $0.52 \pm 0.03$ & $48.50 \pm 3.33$ & $51.17 \pm 2.31$ & $49.98 \pm 3.06$ \\
    \bottomrule
    \toprule
	\textbf{Habitat} & \textbf{SR (\%)} & \textbf{Latency (s/step)} & \textbf{Reuse ratio (\%)} & \textbf{Active ratio (\%)} & \textbf{Diversity ratio (\%)}\\
    \cmidrule(rl){1-1} \cmidrule(rl){2-6}
    \textbf{$\ourmodel$} & $\mathbf{76.27 \pm 3.32}$ & $\mathbf{1.82 \pm 0.03}$ & $\mathbf{72.81 \pm 4.24}$ & $\mathbf{71.38 \pm 3.29}$ & $\mathbf{56.29 \pm 2.82}$ \\
    ReAct & $52.00 \pm 3.32$ & $2.41 \pm 0.07$ & -- & $51.81 \pm 6.33$ & $42.03 \pm 2.12$ \\
    Reflexion & $50.47 \pm 5.40$ & $2.38 \pm 0.18$ & -- & $49.66 \pm 8.71$ & $39.27 \pm 3.04$ \\
    MemRL & $61.80 \pm 2.92$ & $2.68 \pm 0.12$ & -- & $57.55 \pm 4.78$ & $54.36 \pm 1.31$ \\
    CacheBlend & $56.33 \pm 2.24$ & $1.97 \pm 0.03$ & $46.25 \pm 3.29$ & $51.39 \pm 3.29$ & $53.20 \pm 2.30$ \\
    KEEP & $59.40 \pm 2.24$ & $2.09 \pm 0.02$ & $43.05 \pm 4.12$ & $64.58 \pm 1.75$ & $52.40 \pm 1.20$ \\
    LRAgent & $57.47 \pm 3.54$ & $2.03 \pm 0.31$ & $42.96 \pm 3.98$ & $58.34 \pm 4.53$ & $44.28 \pm 3.39$ \\
    \bottomrule
    \end{tabular}
\end{adjustbox}
\end{table*}
\begin{table*}[!b]
\caption{
Performance on real-world embodied tasks.
$\ourmodel$ improves task success and reduces per-step inference time by preserving reusable KV states while using task-relevant memory contents.}
\label{tab:main:real-world}
\centering
\begin{adjustbox}{width=.99\textwidth}
\begin{tabular}{lccccc}
 \toprule
    \textbf{Real-world} & \textbf{SR (\%)} & \textbf{Latency (s/step)} & \textbf{Reuse ratio (\%)} & \textbf{Active ratio (\%)} & \textbf{Diversity ratio (\%)}\\
    \cmidrule(rl){1-1} \cmidrule(rl){2-6}
    $\ourmodel$ & $\mathbf{83.20 \pm 4.39}$ & $\mathbf{1.87 \pm 0.12}$ & $\mathbf{72.18 \pm 3.28}$ & $\mathbf{75.45 \pm 3.92}$ & $\mathbf{64.85 \pm 2.24}$ \\
    MemRL & $64.28 \pm 5.25$ & $2.94 \pm 0.32$ & -- & $61.04 \pm 4.74$ & $58.19 \pm 2.73$ \\
    CacheBlend & $51.21 \pm 4.02$ & $2.14 \pm 0.12$ & $54.02 \pm 4.30$ & $54.27 \pm 3.82$ & $54.37 \pm 4.19$ \\
    KEEP & $63.40 \pm 3.27$ & $2.43 \pm 0.24$ & $48.64 \pm 4.24$ & $62.03 \pm 3.68$ & $56.20 \pm 3.28$ \\
    LRAgent & $53.43 \pm 4.18$ & $2.21 \pm 0.18$ & $50.22 \pm 3.82$ & $56.30 \pm 2.39$ & $49.97 \pm 1.29$ \\
    \bottomrule
    \end{tabular}
\end{adjustbox}
\end{table*}
Table~\ref{tab:main:sim} reports the main results on ALFWorld and Habitat.
Across both benchmarks, $\ourmodel$ achieves the highest task success while also reducing per-step inference time and improving memory utilization.
On ALFWorld, $\ourmodel$ reaches $71.43\%$ SR with $0.18$ seconds per step, outperforming the strongest baseline MemRL at $52.86\%$ SR with $0.90$ seconds per step.
On Habitat, $\ourmodel$ achieves $76.27\%$ SR with $1.82$ seconds per step, improving over MemRL at $61.80\%$ SR with $2.68$ seconds per step.
These gains indicate that reconciling dynamically updated memory contents with reusable KV states provides a more effective and efficient decision context than using episodic traces, reactive prompting, or adapter switching in isolation.

The efficiency and memory behavior metrics explain this trend.
Although CacheBlend, KEEP, and LRAgent reuse KV states, their reuse ratios remain lower than $\ourmodel$ because they do not jointly handle memory-content updates and procedural adapter changes.
CacheBlend relies on non-prefix segments, KEEP invalidates affected memory regions, and LRAgent supports procedural LoRA switching without explicitly processing updates to declarative and working-memory contents.
In contrast, $\ourmodel$ preserves more KV states by indexing memory chunks, updating only affected KV ranges, and maintaining adapter-compatible KV states.
Higher active action ratio and action diversity indicate a higher fraction of actions in task-progressing categories and a higher fraction of unique actions.
Together, these results show that $\ourmodel$ improves long-horizon embodied inference through memory-aware KV reconciliation without rebuilding the full KV cache.

Table~\ref{tab:main:real-world} reports real-world embodied task results.
We focus on MemRL, CacheBlend, KEEP, and LRAgent because they cover the most relevant memory- and cache-specific mechanisms while limiting costly physical robot trials.
$\ourmodel$ achieves the highest SR, outperforming MemRL by $18.92$ percentage points, and reduces inference time from $2.94$ to $1.87$ seconds per step.
This shows that KV reconciliation remains effective when real-world task progress changes action-relevant memory contents.
CacheBlend and LRAgent preserve part of the cache but achieve lower SR, while KEEP improves SR at the cost of lower reuse and higher latency.
By contrast, $\ourmodel$ achieves the highest KV reuse ratio, active action ratio, and action diversity, indicating stronger cache preservation and action-level behavior.

\subsection{Analysis}\label{sec:eval:ana}
\begin{figure*}[!t]
\centering
\includegraphics[width=0.99\linewidth]{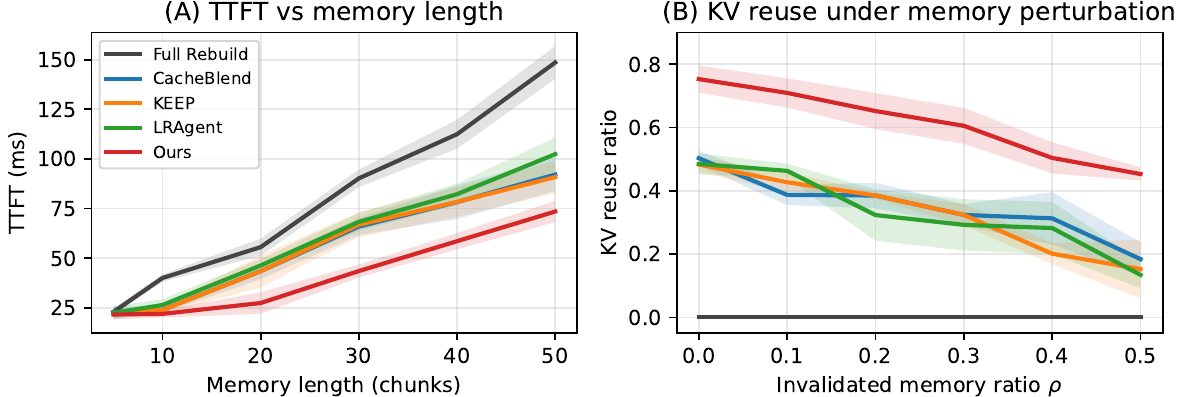}
\caption{
Analysis on Cost efficiency and robustness under memory-content updates.
(A) $\ourmodel$ exhibits slower TTFT growth than the baselines as memory length increases.
(B) $\ourmodel$ preserves higher KV reuse under increasing invalidated memory ratio $\rho$.
}
\label{fig:eval:ana:comb}
\end{figure*}
\paragraph{Cost efficiency under memory-content updates.}
Figure~\ref{fig:eval:ana:comb}(a, b) evaluates cost efficiency as the memory context becomes longer and more frequently invalidated.
Figure~\ref{fig:eval:ana:comb}(a) varies the number of memory chunks in the memory context, while Figure~\ref{fig:eval:ana:comb}(b) varies the invalidated memory ratio $\rho$, defined as the fraction of memory chunks affected between consecutive action decisions.
Full Rebuild shows the steepest TTFT growth because it recomputes the full memory context at every step.
CacheBlend, KEEP, and LRAgent reduce this cost partially, but their gains are limited by narrower reuse assumptions.
CacheBlend depends on reusable non-prefix segments, KEEP relies on invalidating affected memory regions, and LRAgent mainly handles procedural LoRA switching without explicitly reconciling updates to declarative and working-memory contents.

In contrast, $\ourmodel$ achieves the lowest TTFT by localizing memory-content updates to affected KV ranges and preserving matched chunks through RoPE-based remapping.
The reuse-ratio results explain this trend.
Baselines rapidly lose reusable cache regions as the invalidated memory ratio increases, whereas $\ourmodel$ maintains roughly $1.8\times$ higher average reuse and retains more than half of its initial reuse at the largest invalidation ratio.
These results reflect the interdependent roles of our components.
The memory schema identifies affected regions induced by memory-content updates, chunk-level KV reconciliation preserves positional consistency under insertion, eviction, and reordering, and adapter-stable KV reconciliation supports procedural switches.
Together, they enable efficient KV reconciliation without full cache reconstruction.

\begin{table}[b]
\caption{
Analysis on cache-processing cost across reconciliation operations.
$\ourmodel$ recomputes affected KV ranges, whereas Full Rebuild recomputes the full context.}
\label{tab:ana:operation}
\centering
\begin{adjustbox}{max width=.99\textwidth}
\begin{tabular}{ll rrrrrr r}
\toprule
& & \textbf{Reuse} & \textbf{Reorder} & \textbf{Recompute} & \textbf{Insert} & \textbf{Evict} & \textbf{Assemble $\mathcal{K}_t$} & \textbf{Total} \\
\midrule
\multirow{3}{*}{$\ourmodel$}
 & Events   & 5,167  & 528   & 418   & 408   & 113   & --  & 6,634 \\
 & Tokens   & 64,128 & 6,437 & 4,219 & 5,358 & 1,742 & --  & 81,884 \\
 & Time (s) & 0.0    & 0.0   & 12.9  & 12.4  & 0.0   & 0.5 & $\mathbf{25.8}$ \\
\midrule
\multirow{3}{*}{Full Rebuild}
 & Events   & 0  & 0  & 6,113  & 408   & 113   & --  & 6,634 \\
 & Tokens   & 0  & 0  & 74,784 & 5,358 & 1,742 & --  & 81,884 \\
 & Time (s) & -- & -- & 186.5  & 12.4  & 0.0   & 0.5 & 199.4 \\
\bottomrule
\end{tabular}
\end{adjustbox}
\end{table}

\paragraph{Operation-level cost decomposition.}
Table~\ref{tab:ana:operation} decomposes cache-processing cost by reconciliation operation, replaying the sequences of memory-content updates collected from ALFWorld episodes, where the reconciliation plan $\Delta_t$ maps each update to insertion, eviction, reordering, recomputation, or reuse.
Most of the memory context remains reusable across decision steps: $78.3\%$ of the processed tokens are reused and $7.9\%$ are relocated with RoPE-based remapping, both without recomputation, while only $4{,}219$ tokens, $5.6\%$ of the $74{,}784$ tokens that Full Rebuild recomputes over the same updates, are recomputed.
Because insertion and eviction costs are identical to Full Rebuild by construction, the entire reduction in cache-processing time, from $199.4$ to $25.8$ seconds, comes from recomputation avoided on schema-compatible KV ranges.
Assembling the reconciled KV cache $K_t$ adds only 0.5 seconds in total, indicating low cache-assembly overhead relative to recomputation.
Note that the reuse ratio in Table~\ref{tab:main:sim} is a per-step average over episodes, whereas the token counts here aggregate all replayed memory-content updates.

Further analysis in Appendix~\ref{app:analysis:scaling} evaluates scalability under increasing memory chunk counts and dependency density, as well as the memory footprint and switching overhead of procedural adapters.
Appendix~\ref{app:analysis:latency} further decomposes per-step inference latency, showing that the reduction in local memory-context recomputation accounts for most of the observed TTFT improvement.

\subsection{Ablation}\label{sec:eval:abl}\label{sec:ablation}

\begin{table*}[t]
\centering
\caption{Ablation on $\ourmodel$. ‘w/o.’ denotes the removal of a component.}
\label{tab:ablation}

\resizebox{\textwidth}{!}{%
\begin{tabular}{lccccc}
\toprule
\textbf{Variant} & \textbf{SR (\%)} & \textbf{Latency (s/step)} & \textbf{Reuse ratio(\%)} & \textbf{Active ratio (\%)} & \textbf{Diversity ratio(\%)}\\
\midrule
\textbf{Ours} & $\mathbf{71.43 \pm 2.05}$ & $\mathbf{0.18 \pm 0.02}$ & $\mathbf{75.30 \pm 4.33}$ & $\mathbf{67.69 \pm 2.27}$ & $\mathbf{71.43 \pm 3.32}$ \\
w/o. Memory Schema & $62.67 \pm 3.49$ & $0.38 \pm 0.12$ & $51.39 \pm 5.84$ & $57.20 \pm 4.12$ & $61.00 \pm 5.03$ \\
w/o. Chunk-Level KV Reconciliation & $59.40 \pm 4.27$ & $0.47 \pm 0.05$ & $42.75 \pm 6.20$ & $55.10 \pm 4.75$ & $59.25 \pm 5.61$ \\
w/o. Adapter-Stable KV Reconciliation & $64.85 \pm 3.58$ & $0.28 \pm 0.03$ & $66.40 \pm 4.92$ & $62.71 \pm 3.66$ & $65.15 \pm 4.44$ \\
\bottomrule
\end{tabular}%
}
\end{table*}
Table~\ref{tab:ablation} ablates the main components of $\ourmodel$.
Removing the memory schema decreases SR from $71.43\%$ to $62.67\%$ and reuse from $75.30\%$ to $51.39\%$.
This shows that anchoring memory contents as typed chunks with KV ranges is necessary for identifying which contents remain reusable after memory-content updates.
Without this schema, the model cannot reliably map memory contents to affected KV ranges, leading to lower reuse, higher latency, and a lower active action ratio.

Removing chunk-level KV reconciliation causes the largest degradation, reducing SR to $59.40\%$ and reuse to $42.75\%$, while increasing latency to $0.47$ seconds per step.
This indicates that the reconciliation plan is the main mechanism for localizing memory-content updates.
When chunk-level reconciliation is removed, insertions, evictions, and reorderings can no longer be handled by updating only affected KV ranges, forcing broader recomputation and making cache reuse less stable.

Removing adapter-stable KV reconciliation also degrades performance, but differently.
The reuse ratio remains high at $66.40\%$, yet SR drops to $64.85\%$ and latency increases to $0.28$ seconds per step.
This suggests that reusing many KV states is insufficient when procedural memory changes the active adapter.
Without adapter-stable reconciliation, KV states computed under a previous adapter can become incompatible with the current procedural configuration, reducing decision quality even when cache reuse is preserved.
Overall, the ablation results show that the memory schema identifies reusable contents, chunk-level reconciliation localizes memory-content updates, and adapter-stable reconciliation ensures reused states remain compatible under procedural changes.

\section{Conclusion}\label{sec:con}
We presented $\ourmodel$, a heterogeneous memory reconciliation framework for foundation model-based embodied agents.
$\ourmodel$ anchors memory contents as typed memory chunks with associated KV ranges and reconciles memory-content updates through localized KV operations.
By combining chunk-level and adapter-stable KV reconciliation, it preserves compatible KV states across changes to declarative, procedural, and working memory contents without rebuilding the full KV cache.
Experiments on ALFWorld, Habitat, and real-world tabletop manipulation tasks show that $\ourmodel$ improves task success and reduces per-step inference time over strong baselines.
These results demonstrate that reconciling dynamically updated memory contexts with reusable KV states supports efficient and stable inference in long-horizon embodied tasks.

\paragraph{Limitation and future work.}
Our evaluation focuses on single-agent embodied tasks, where memory contexts are updated across successive decision steps.
In multi-agent embodied systems, concurrent updates to shared memory contents may introduce consistency issues between memory contexts and cached KV states, which are not addressed by the current framework.
As future work, we plan to extend $\ourmodel$ to multi-agent embodied systems by incorporating concurrency control for shared memory updates and KV reuse.
This extension may support efficient memory reconciliation across agents while preserving valid KV states under concurrent updates.

\newpage
\section*{Acknowledgment}
This work was supported
by the Institute of Information \& Communications Technology Planning \& Evaluation (IITP) grants funded by the Korea government (MSIT)
(RS-2025-02218768, Accelerated Insight Reasoning via Continual Learning (70\%);
RS-2025-25442569, AI Star Fellowship Support Program (Sungkyunkwan Univ.) (10\%)),
by the National Research Foundation of Korea (NRF) grant funded by the Korea government (MSIT) (RS-2026-25474409, 10\%),
and by Samsung Electronics Co., Ltd. (10\%).
\printbibliography

\clearpage
\newpage
\appendix
\onecolumn
\section{Experiment setting}\label{app:exp}
\subsection{ALFWorld}
\label{app:exp:alfworld}

\begin{figure}[h]
    \centering
    \renewcommand{\arraystretch}{0.85} 
    \begin{minipage}[b]{0.47\linewidth}
        \centering
        \includegraphics[width=\linewidth, trim=0 9mm 0 0, clip]{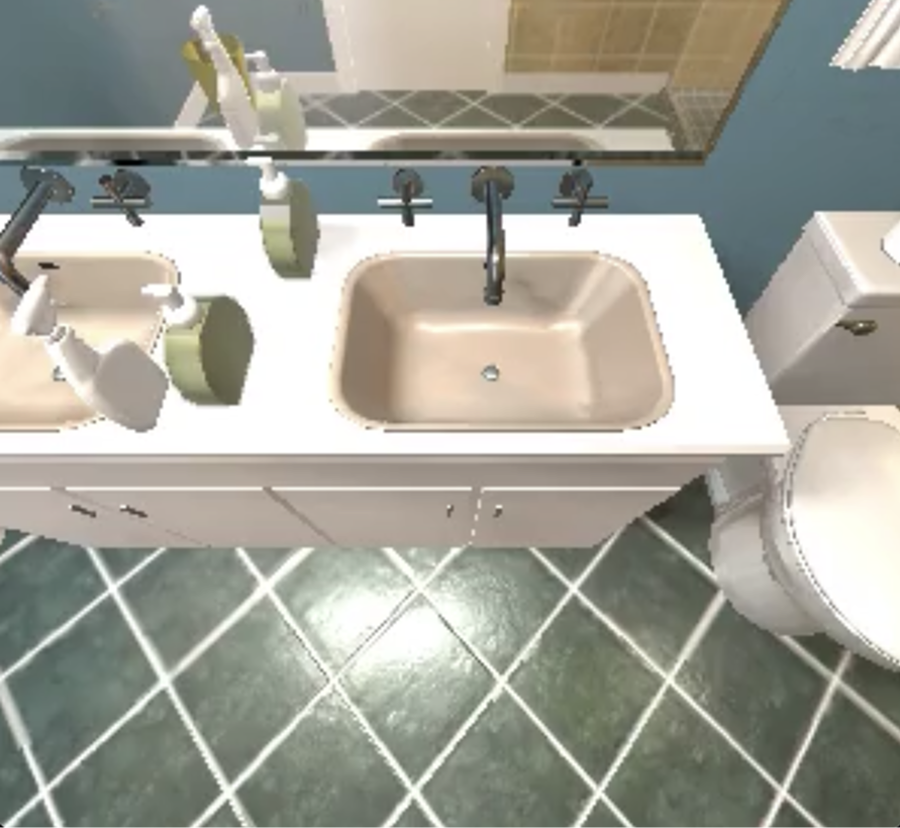}
        \centerline{(a) Bathroom scene example}
    \end{minipage}
    \hfill
    \begin{minipage}[b]{0.47\linewidth}
        \centering
        \includegraphics[width=\linewidth, trim=0 9mm 0 0, clip]{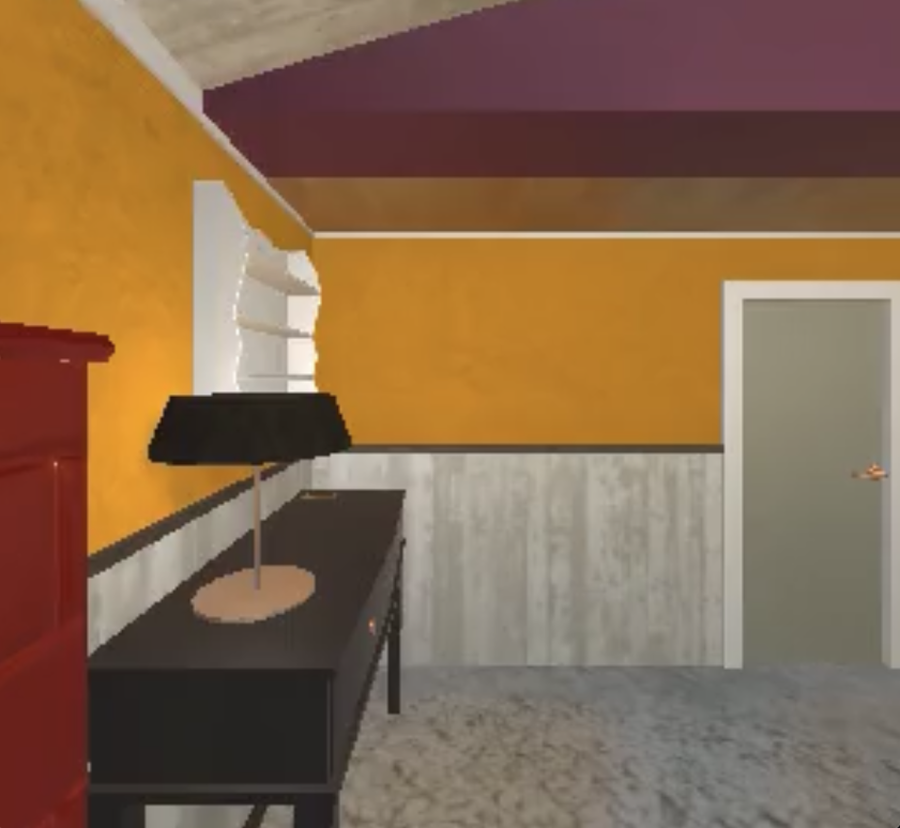}
        \centerline{(b) Bedroom scene example}
    \end{minipage}
    
    \vspace{2mm}
    
    \begin{minipage}[b]{0.47\linewidth}
        \centering
        \includegraphics[width=\linewidth]{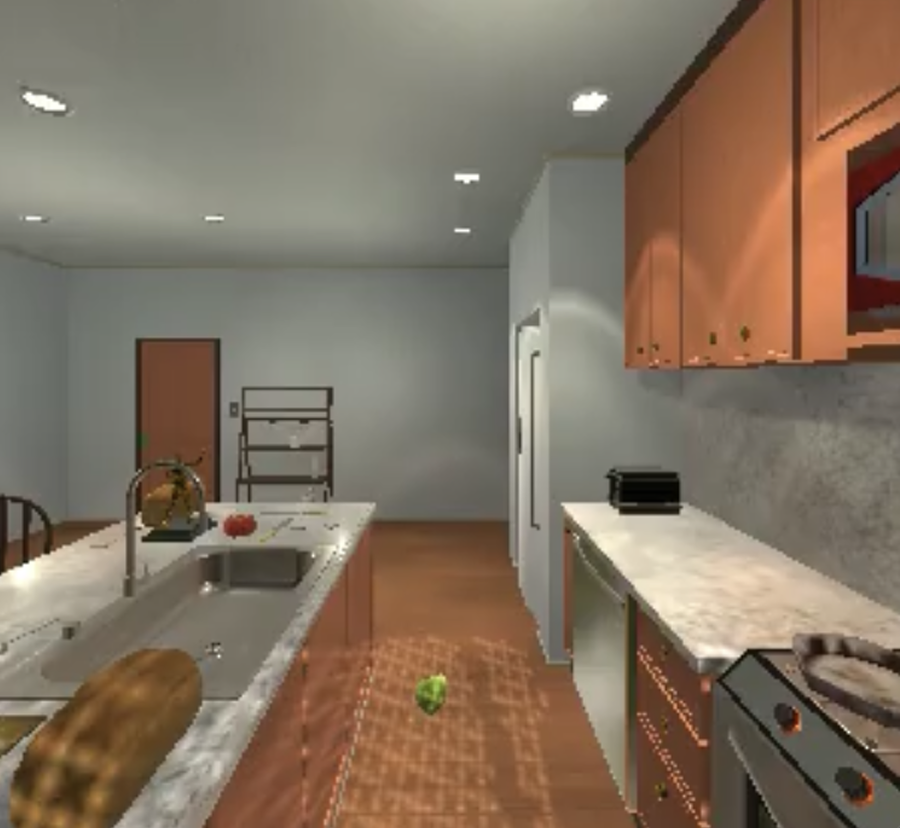}
        \centerline{(c) Kitchen scene example}
    \end{minipage}
    \hfill
    \begin{minipage}[b]{0.47\linewidth}
        \centering
        \includegraphics[width=\linewidth]{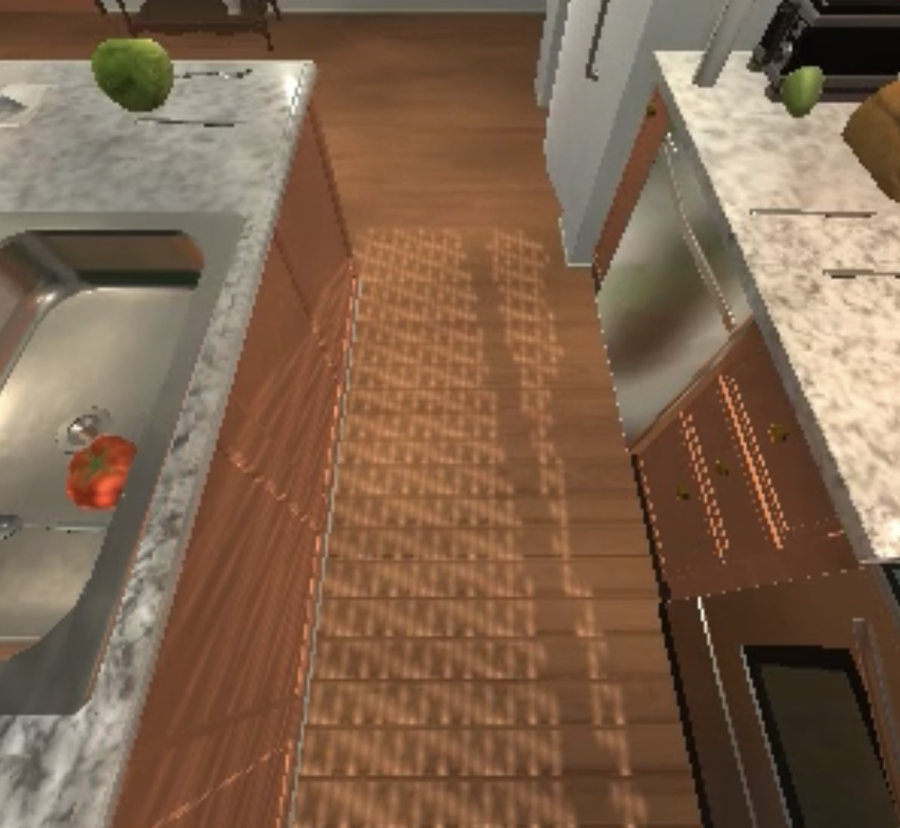}
        \centerline{(d) Kitchen scene example}
    \end{minipage}
    \caption{Example ALFWorld scenes.}
    \label{app:fig:ben:alfworld}
\end{figure}

\paragraph{Environment.\quad}
ALFWorld~\cite{ben:alfworld} is a text-based household environment in which an agent solves natural-language instructions through a discrete language action interface, as shown in Figure~\ref{app:fig:ben:alfworld}.
It inherits the task structure of ALFRED, exposing textual observations and an admissible-command list at every decision step.
Each episode starts from the initial observation produced by the TextWorld backend.
At every step, the agent receives a textual observation, an admissible-command list with about 30 commands on average, and selects a single command.
We cap each episode at 30 steps, after which the episode is marked as failed if the goal predicate is not satisfied.
Across all baselines and our method, observations are normalized through the same name-canonicalization layer.
For example, \texttt{desk\_bar\_\_+01\_dot\_49\_\_\dots} is rewritten to \texttt{desk\_1}, so prompt content depends on the agent's behavior rather than positional artifacts in receptacle names. ALFWorld is released under the MIT License, and we used it in accordance with the terms specified.

\paragraph{Task.\quad}
Each episode is one of six task types: \emph{put}, \emph{put-two}, \emph{heat}, \emph{cool}, \emph{clean}, and \emph{examine}.
The instruction templates and text action grammar used in the ALFWorld
experiments are summarized in Table~\ref{app:tab:ben:alfworld}.
Procedural memory uses room-specific LoRA adapters for bathrooms, bedrooms, and living rooms, together with a shared adapter.
The agent selects the adapter according to the room context and uses the shared adapter when no room-specific adapter applies.
ALFWorld therefore provides a text-only setting in which memory-content updates arise through changing observations, admissible actions, object locations, and subgoal progress.

\begin{table}[!h]
\caption{Instructions and actions in ALFWorld experiment.}
\label{app:tab:ben:alfworld}
\begin{center}
\begin{adjustbox}{width=1.\textwidth}
\begin{tabular}{l ll}
\toprule
& \textbf{Template} & \textbf{Example} \\
\midrule
\multirow{6}{*}{Instruction}  
                        & Put Object
                            & \textit{``Put the apple on the dining table.''} \\
                        & Put Two Objects
                            & \textit{``Put two pens in the drawer.''} \\
                        & Heat Object
                            & \textit{``Heat the potato with the microwave.''} \\
                        & Cool Object
                            & \textit{``Cool the lettuce with the fridge.''} \\
                        & Clean Object
                            & \textit{``Clean the mug with the sink.''} \\
                        & Examine Object
                            & \textit{``Examine the lamp with the desk lamp.''} \\

\midrule
\multirow{9}{*}{Text Action Grammar} 
                        & Go to Receptacle
                            & \texttt{go to recep} \\ 
                        & Take Object
                            & \texttt{take object from recep} \\
                        & Move Object
                            & \texttt{move object to recep} \\
                        & Open Receptacle
                            & \texttt{open recep} \\
                        & Close Receptacle
                            & \texttt{close recep} \\
                        & Use Object
                            & \texttt{use object} \\
                        & Clean Object
                            & \texttt{clean object with recep} \\
                        & Heat Object
                            & \texttt{heat object with recep} \\
                        & Cool Object
                            & \texttt{cool object with recep} \\
\bottomrule
\end{tabular}
\end{adjustbox}
\end{center}
\end{table}

\subsection{EmbodiedBench Habitat}
\label{app:exp:habitat}

\begin{figure}[h]
    \centering
    \begin{minipage}[b]{0.80\linewidth}
        \centering
        \includegraphics[width=\linewidth, trim=0 9mm 0 0, clip]{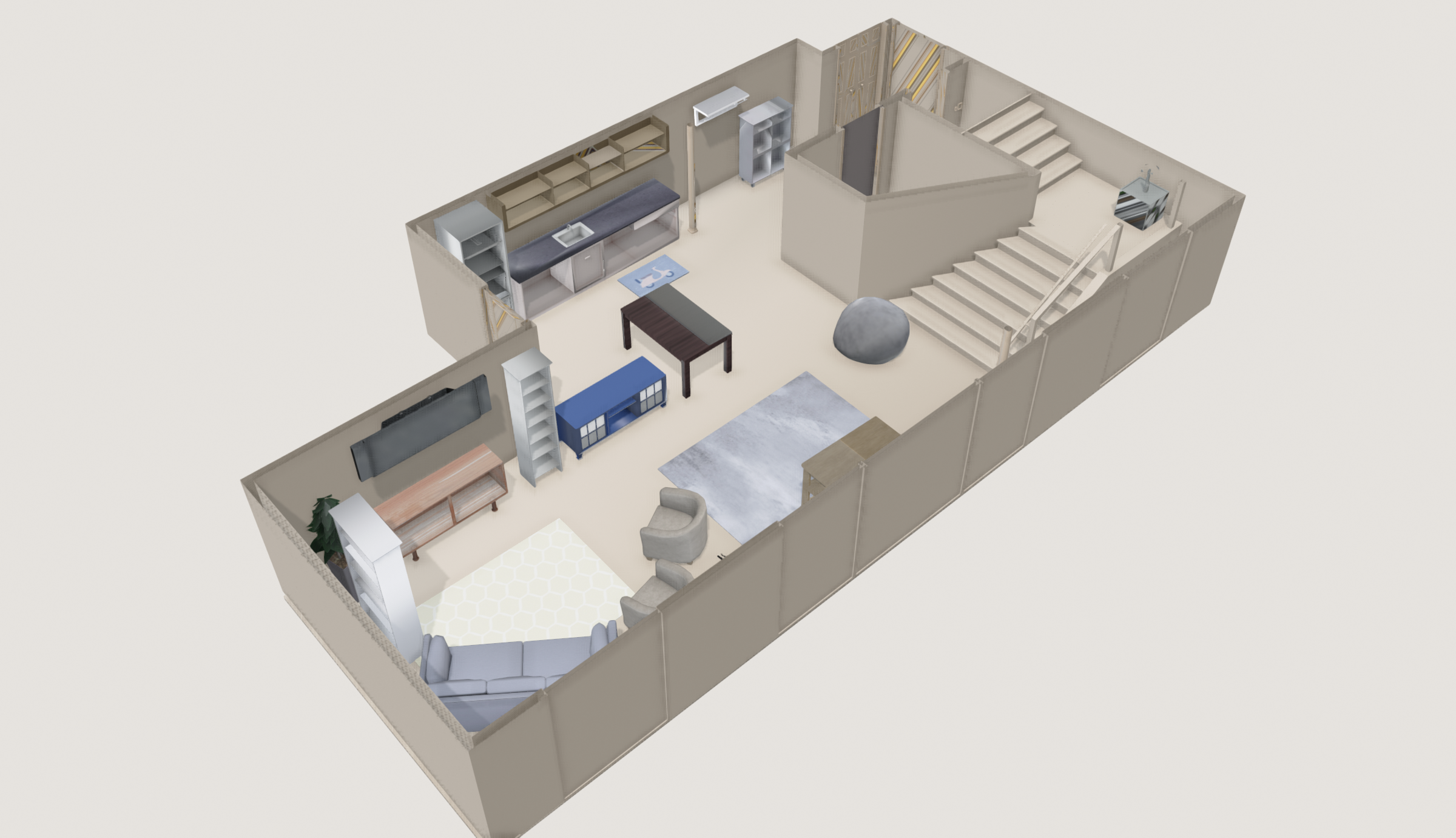}
        \centerline{(a) Top-view image of an EmbodiedBench Habitat environment}
    \end{minipage}
    
    \vspace{2mm}
    
    \begin{minipage}[b]{0.27\linewidth}
        \centering
        \includegraphics[width=\linewidth]{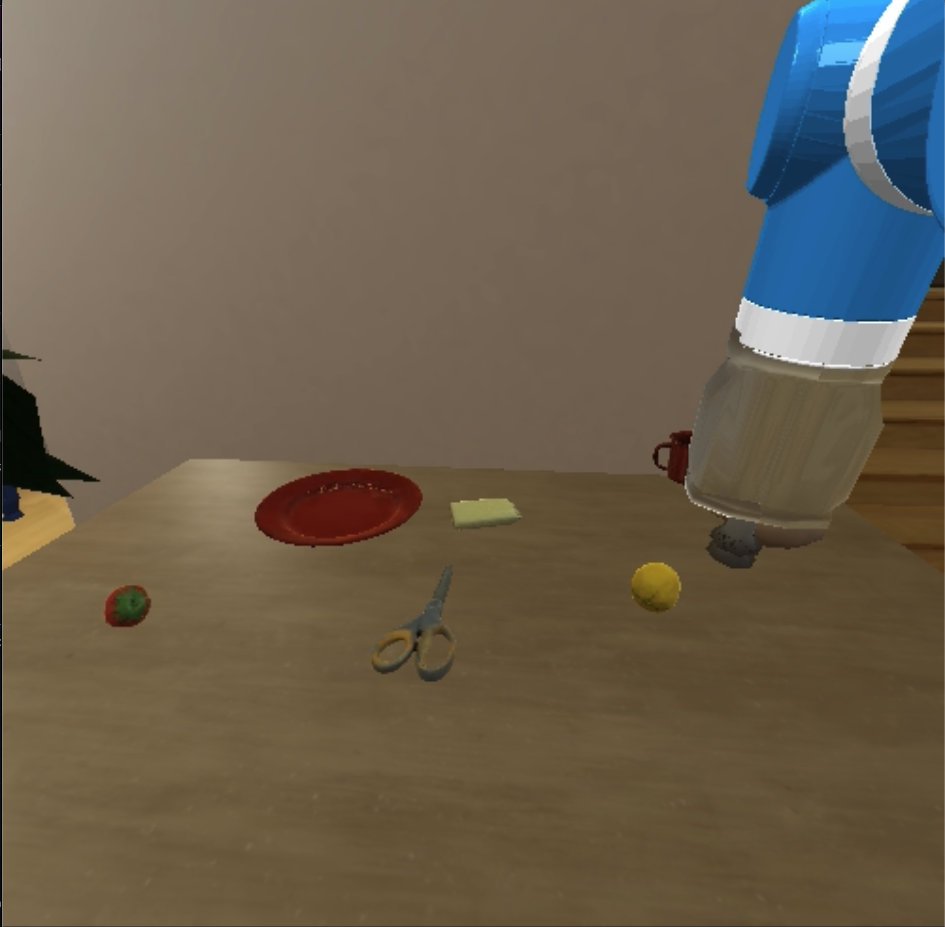}
    \end{minipage}
    \begin{minipage}[b]{0.27\linewidth}
        \centering
        \includegraphics[width=\linewidth]{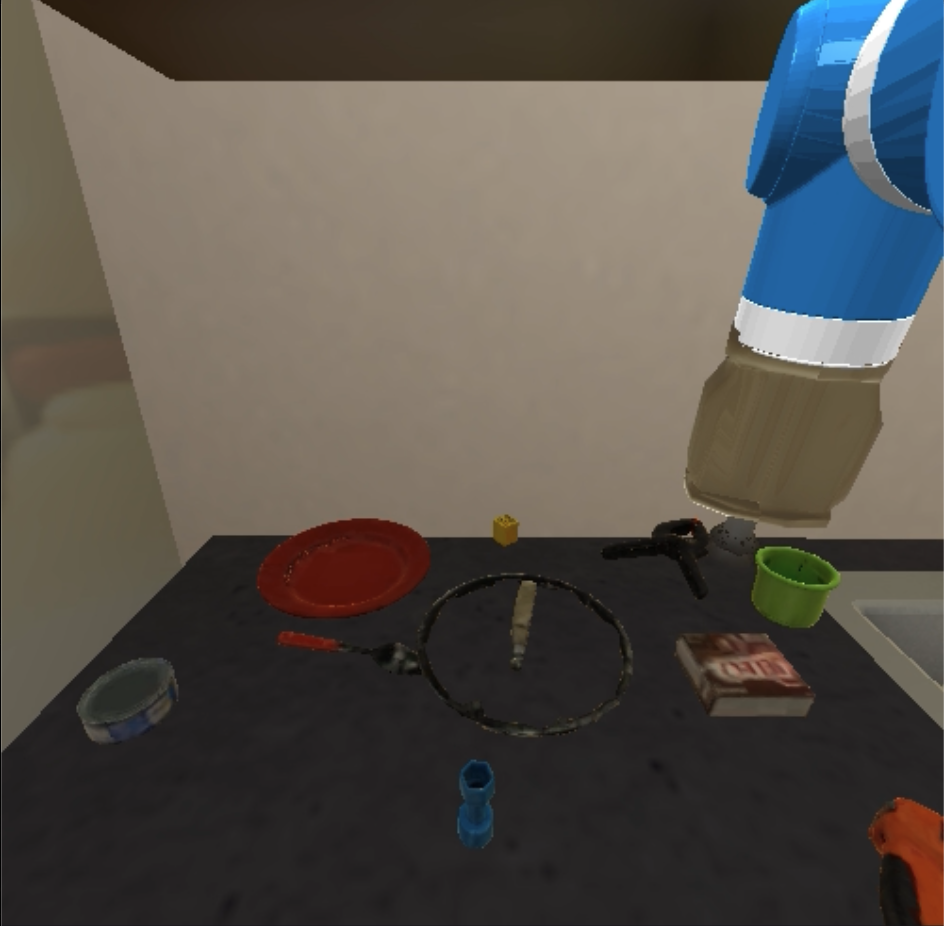}
    \end{minipage}
    \begin{minipage}[b]{0.27\linewidth}
        \centering
        \includegraphics[width=\linewidth]{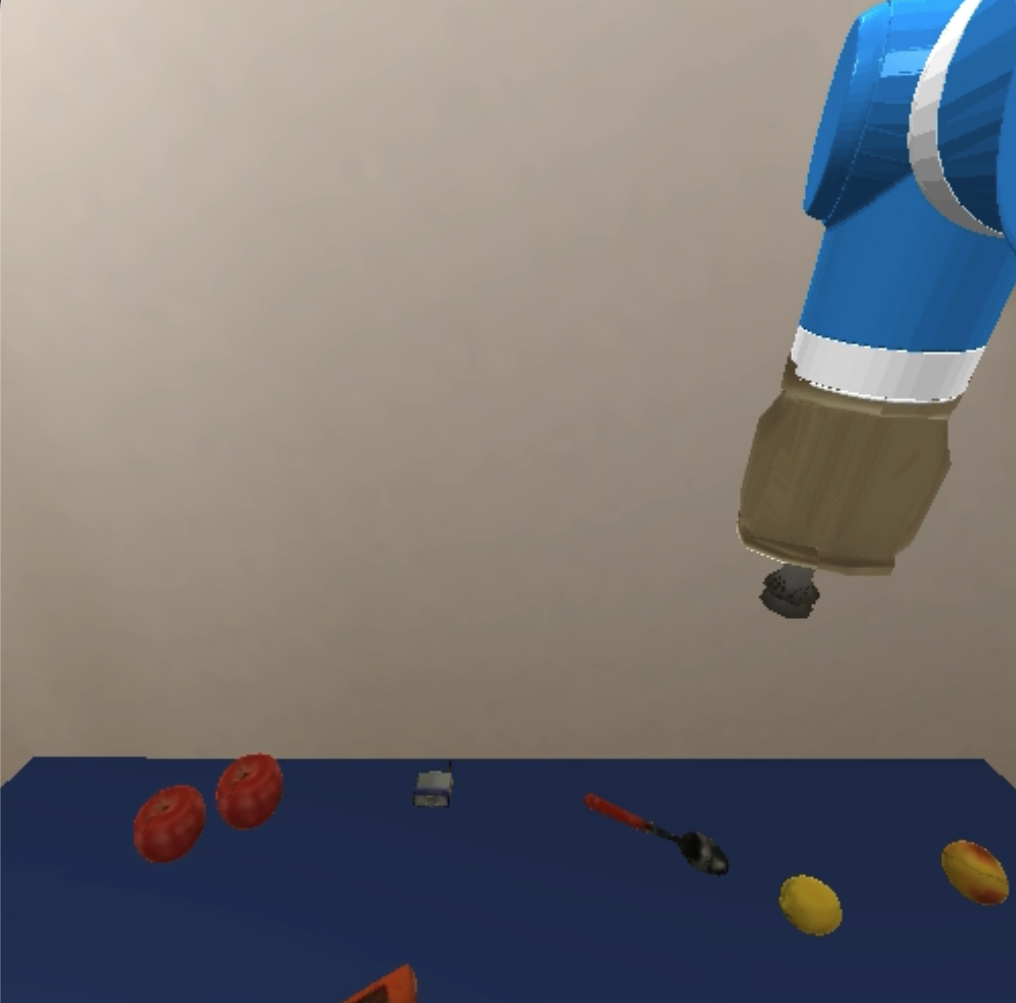}
    \end{minipage}
    \centerline{(b) Task example: ``deposit apples on tv stand''}
    \caption{Example EmbodiedBench Habitat scene and task observations.}
    \label{app:fig:ben:habitat}
\end{figure}

\paragraph{Environment.\quad}
EmbodiedBench Habitat~\cite{ben:embodiedbench,ben:habitat} shown in Figure~\ref{app:fig:ben:habitat} exposes ReplicaCAD apartment scenes through a high-level skill action space.
Unlike ALFWorld, Habitat provides RGB observations, so the agent must condition on visual input in addition to language goals.
In our evaluation, episodes are constructed from six ReplicaCAD apartments, \texttt{apt\_0}--\texttt{apt\_5}, with randomized object placements.
Visual observations are rendered as $512{\times}512$ head-mounted RGB images and downsampled to the resolution accepted by the backbone vision encoder.
The agent acts through high-level skills: \texttt{navigate\_to(receptacle\_or\_object)}, \texttt{pick(object)}, \texttt{place(receptacle)}, \texttt{open(receptacle)}, and \texttt{close(receptacle)}, as
summarized in Table~\ref{app:tab:ben:habitat}.
Each skill is internally executed by a low-level controller, while the high-level decision step is the unit at which $\ourmodel$ reconciles memory chunks.
Each episode is capped at 50 high-level steps. 
Habitat-Sim and Habitat-Lab are released under the MIT License, and we used them in accordance with the terms specified in their official repositories.
ReplicaCAD scene assets are released under the Creative Commons Attribution 4.0 International (CC BY 4.0) License, and we used them in accordance with its terms.

\paragraph{Task.\quad}
EmbodiedBench Habitat includes six evaluation sets that emphasize different aspects of compositional reasoning: \emph{base}, \emph{common sense}, \emph{complex instruction}, \emph{spatial relationship}, \emph{visual appearance}, and \emph{long horizon}.
Each set contains 50 episodes, and we report aggregate scores across all six sets in the main results.
Habitat provides a complementary multimodal setting where memory-content updates arise from both visual observations and high-level skill transitions.
The apartment layout supplies persistent declarative structure, while object visibility, object state, and manipulation outcomes change across decision steps.
This produces a mixture of stable memory contents, newly grounded visual facts, and updated working-memory states.
The high-level skill interface also makes procedural boundaries explicit: switching from navigation to manipulation, or from opening to placing, changes the procedural context that determines the active adapter.
For multimodal backbones, $\ourmodel$ restricts RoPE-based remapping to text-side memory chunks and encodes current visual observations normally at each decision step.

\begin{table}[!h]
\caption{Instructions and skill APIs in EmbodiedBench Habitat experiment.}
\label{app:tab:ben:habitat}
\begin{center}
\begin{adjustbox}{width=1.\textwidth}
\begin{tabular}{l ll}
\toprule
& \textbf{Template} & \textbf{Example} \\
\midrule
\multirow{6}{*}{Instruction}  
                        & Base Task
                            & \textit{``Put the apple on the table.''} \\
                        & Common-sense Task
                            & \textit{``Place the mug near the coffee machine.''} \\
                        & Complex Instruction
                            & \textit{``Pick up the book, open the cabinet, and place it inside.''} \\
                        & Spatial Relationship
                            & \textit{``Put the bowl to the left of the plate.''} \\
                        & Visual Appearance
                            & \textit{``Pick up the red object on the counter.''} \\
                        & Long-horizon Task
                            & \textit{``Move the apple from the kitchen counter to the TV stand.''} \\

\midrule
\multirow{5}{*}{High-level Skill APIs} 
                        & Navigate to Object or Receptacle
                            & \texttt{navigate\_to(receptacle\_or\_object)} \\ 
                        & Pick Object
                            & \texttt{pick(object)} \\
                        & Place Object
                            & \texttt{place(receptacle)} \\
                        & Open Receptacle
                            & \texttt{open(receptacle)} \\
                        & Close Receptacle
                            & \texttt{close(receptacle)} \\
\bottomrule
\end{tabular}
\end{adjustbox}
\end{center}
\end{table}

\subsection{Real-world Test}
\begin{figure}[h]
    \centering
    \begin{minipage}[b]{\linewidth}
        \centering
        \begin{minipage}[c]{0.6\linewidth}
            \centering
            \includegraphics[width=\linewidth]{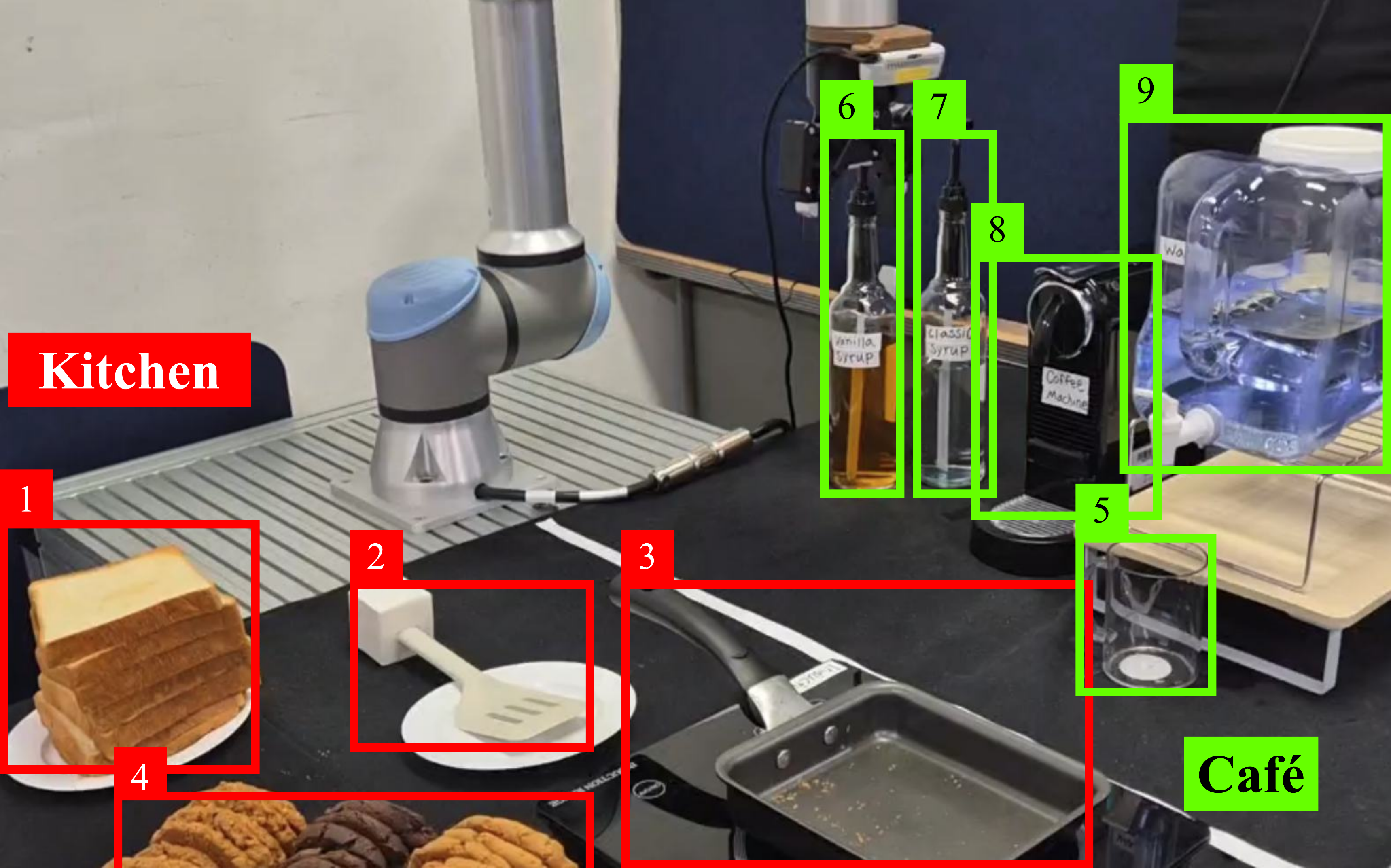}
        \end{minipage}
        \hfill
        \begin{minipage}[c]{0.34\linewidth}
            \centering
            \small
            \begin{tabular}{c|l}
                \hline
                \textbf{ID} & \textbf{Object} \\
                \hline
                1 & Bread slices \\
                2 & Spatula \\
                3 & Frying pan \\
                4 & Cookie \\
                5 & Cup \\
                6 & Vanilla syrup bottle \\
                7 & Classic syrup bottle \\
                8 & Coffee machine \\
                9 & Water dispenser \\
                \hline
            \end{tabular}
        \end{minipage}
        \centerline{(a) Example of \textit{Dual-Section Kitchen-Caf\'e} environment}
    \end{minipage}
    
    \vspace{3mm}
    
        \begin{minipage}[b]{\linewidth}
        \centering
        \begin{minipage}[c]{0.6\linewidth}
            \centering
            \includegraphics[width=\linewidth]{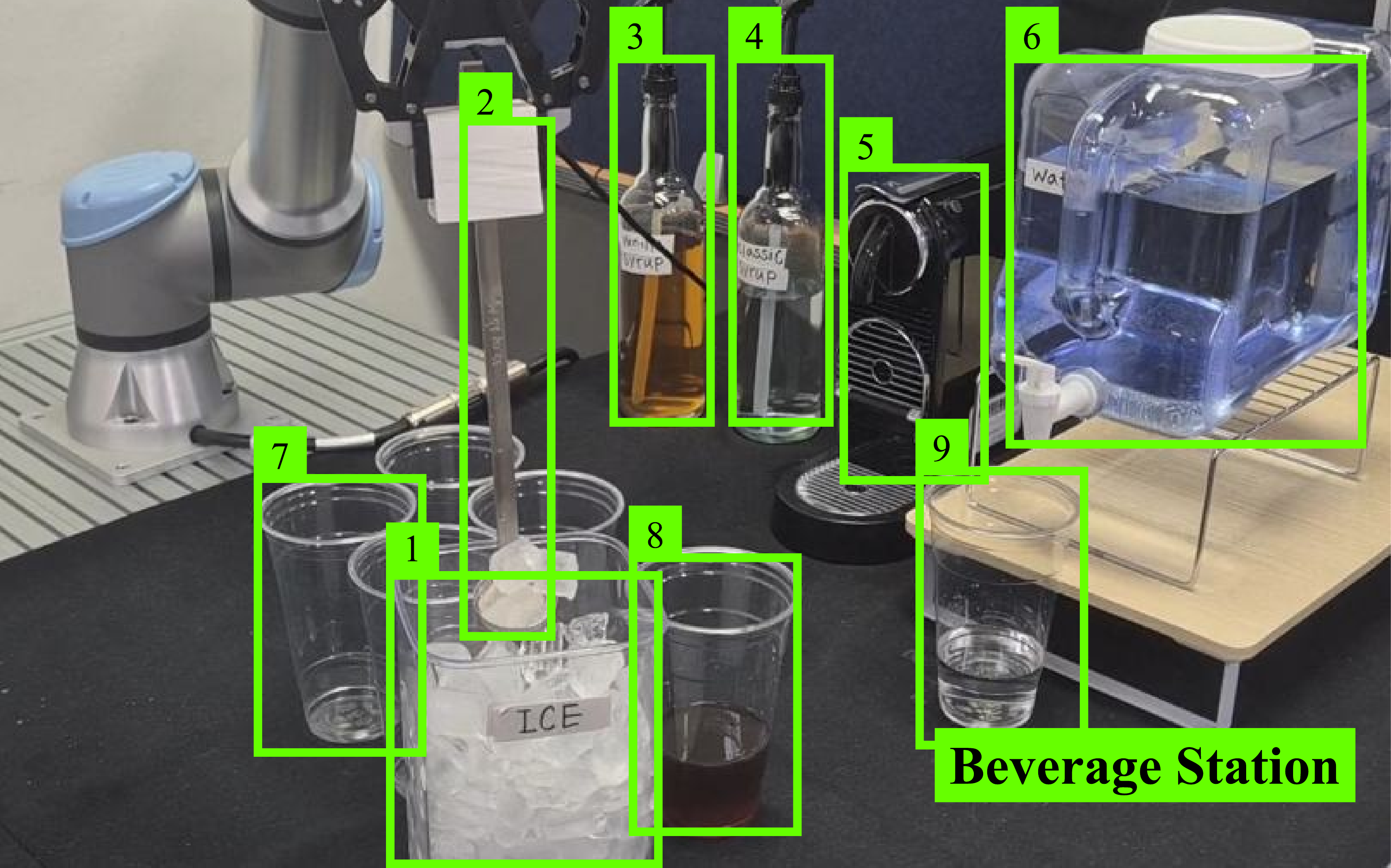}
        \end{minipage}
        \hfill
        \begin{minipage}[c]{0.34\linewidth}
            \centering
            \small
            \begin{tabular}{c|l}
                \hline
                \textbf{ID} & \textbf{Object} \\
                \hline
                1 & Ice container \\
                2 & Ice scoop \\
                3 & Vanilla syrup bottle \\
                4 & Classic syrup bottle \\
                5 & Coffee machine \\
                6 & Water dispenser \\
                7 & cup \\
                8 & coffee cup \\
                9 & water cup \\
                \hline
            \end{tabular}
        \end{minipage}
        \centerline{(b) Example of \textit{Single-Section Beverage Station set} environment}
    \end{minipage}
    
    \caption{Example real-world environments.}
    \label{app:fig:ben:realworld}
\end{figure}
\paragraph{Environment.\quad}
Our real-world setup employed a UR7e robotic arm mounted on a tabletop workspace. For perception, we used two Intel RealSense D435 RGB-D cameras: one attached to the gripper for close-range, viewpoint-aligned observation during manipulation, and another mounted overhead in a bird's-eye configuration to capture the global scene layout. The two depth streams were fused into a unified point cloud, yielding a high-resolution 3D reconstruction of the workspace that served as the basis for object localization and visual grounding. Perception and control were integrated in a single real-time loop, ensuring consistent behavior throughout manipulation.

\paragraph{Task.\quad}
Real-world tasks are executed using a compact set of predefined task-level skill APIs. As summarized in Table~\ref{app:tab:ben:realworld}, these APIs cover common tabletop manipulation operations, including pose-conditioned motion, object picking, object placement, pick-and-place transfer, directional pushing, tool grasping, and object flipping. Each skill exposes only the task-relevant arguments needed for planning, such as the target object, source and target locations, end-effector pose, or push direction. This abstraction hides low-level control details while preserving sufficient flexibility for the LLM-based agent to compose actions across cooking and beverage-preparation tasks.

Real-world experiments were conducted in two tabletop environments: a Dual-Section Kitchen-Caf\'e environment and a Single-Section Beverage Station environment.
Each of the two evaluated long-horizon tasks is repeated with varying object configurations.
At the beginning of each trial, objects were sampled from the corresponding object pool and placed at randomized positions within the workspace.
We repeat each task across multiple randomized physical trials ($N=5$ per task) to evaluate robustness under varied instructions and spatial layouts.
Simulator results are averaged over five seeds, while real-world results are averaged over these physical trials.

The first environment is a Dual-Section Kitchen-Caf\'e setup, which combines a kitchen section and a caf\'e section within a single tabletop workspace. The kitchen section includes objects such as bread slices, a spatula, and a frying pan, while the caf\'e section includes a cup, syrup bottles, a coffee machine, and a water dispenser. Tasks in this environment require the agent to compose skills across the two sections, such as placing bread on the pan, flipping toast, moving a cup to the appropriate station, and operating beverage-related devices. This setup is designed to evaluate whether $\ourmodel$ can compose procedural knowledge across heterogeneous but co-located task domains, while reusing shared manipulation skills such as picking, placing, pushing, and tool grasping.

The second environment is a Single-Section Beverage Station setup, which focuses only on the caf\'e-side task domain. It contains beverage-related objects such as an ice container, ice scoop, syrup bottles, a coffee machine, a water dispenser, and a cup. Tasks in this environment focus on beverage preparation, including placing the cup, checking the state of the ice, adding ingredients, operating the dispenser or coffee machine through directional pushes, and manipulating tools when needed. Compared with the dual-section setup, this environment isolates a single procedural domain and is used to evaluate whether $\ourmodel$ can efficiently execute and reuse a focused set of beverage-preparation skills while adapting to local state changes, such as whether ice is available or needs to be added. Example images of both environments are shown in Figure~\ref{app:fig:ben:realworld}.

\begin{table}[!h]
\caption{Instructions and APIs in Real-world Experiment}
\label{app:tab:ben:realworld}
\begin{center}
\begin{adjustbox}{width=1.\textwidth}
\begin{tabular}{l ll}
\toprule
& \textbf{Template} & \textbf{Example} \\
\midrule
\multirow{7}{*}{Instruction}  
                        & Pick Object
                            & \textit{``Pick up the cup from the table.''} \\
                        & Pick \& Place
                            & \textit{``Pick up a bread slice and place it on the frying pan.''} \\
                        & Place Object
                            & \textit{``Place the cup under the water dispenser.''} \\
                        & Push
                            & \textit{``Push the dispenser lever downward.''} \\
                        & Grasp Tool
                            & \textit{``Grasp the spatula.''} \\
                        & Flip Object
                            & \textit{``Flip the toast on the frying pan.''} \\
                        & Serve Beverage
                            & \textit{``Prepare a drink using water, syrup, and espresso.''} \\

\midrule
\multirow{7}{*}{Task-level Skill APIs} 
                        & Move to Pose [x,y,z,r,p,y]
                            & \texttt{pose(x, y, z, r, p, y)} \\ 
                        & Pick [object, location]
                            & \texttt{pick(object, location)} \\
                        & Place [object, location]
                            & \texttt{place(object, location)} \\
                        & Pick \& Place [object, start, target]
                            & \texttt{pick\_and\_place(object, start\_location, target\_location)} \\
                        & Push [location, direction]
                            & \texttt{push(location, direction)} \\
                        & Grasp Tool [tool]
                            & \texttt{grasp\_tool(tool)} \\
                        & Flip [object]
                            & \texttt{flip(object)} \\
\bottomrule
\end{tabular}
\end{adjustbox}
\end{center}
\end{table}

\subsection{Real-world experimental details}
\subsubsection{Dual-Section Kitchen-Caf\'e}
\begin{figure}[!h]
    \centering
    \includegraphics[width=0.8\linewidth]{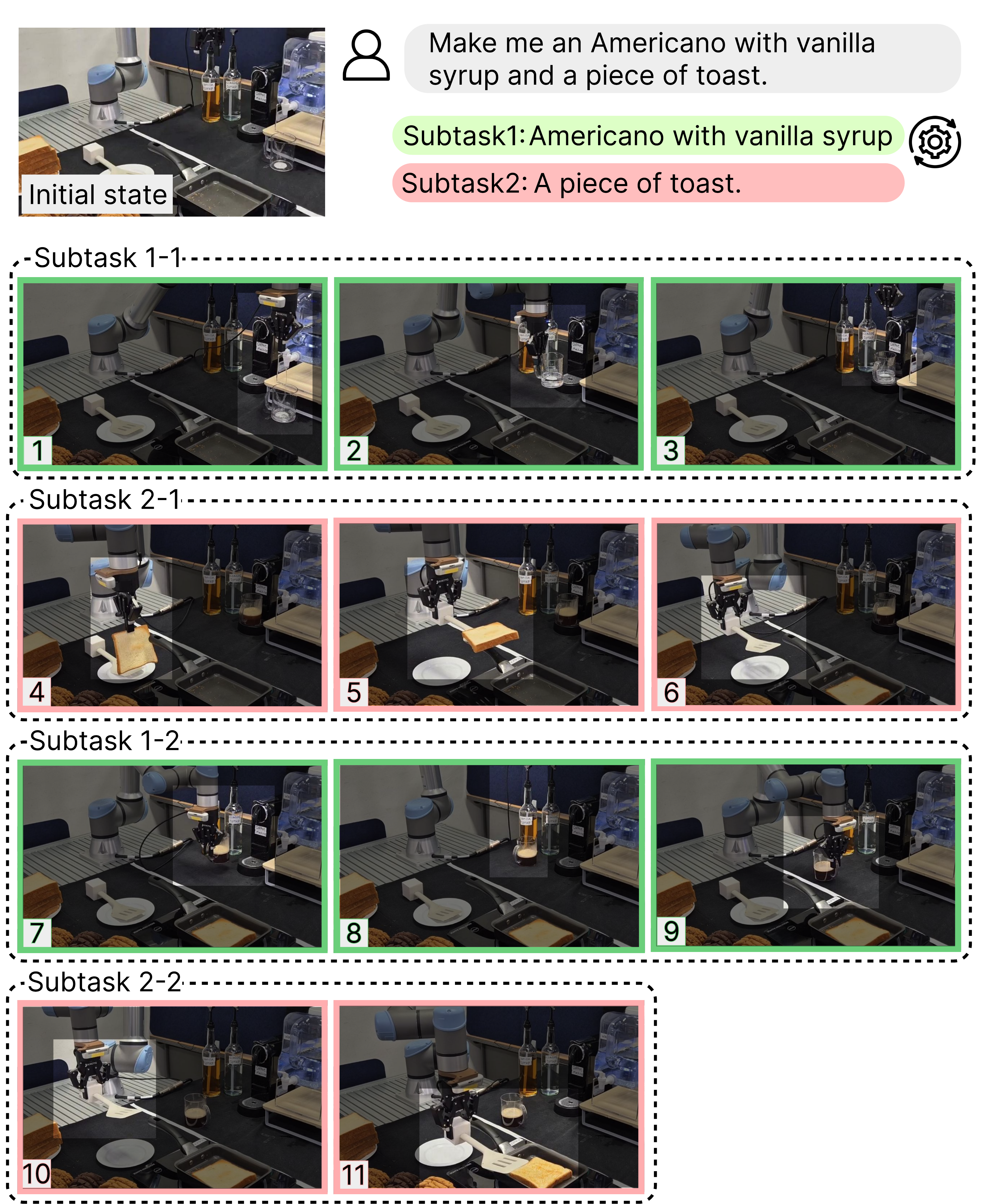}
    \caption{Real-world \textit{Dual-Section Kitchen-Caf\'e}. The sequence includes two main subtasks: (1) making coffee with vanilla syrup and (2) making toast.}
    \label{app:fig:ben:realworld-rollout-1}
\end{figure}

The first real-world environment includes a \textit{dual-section kitchen-café} environment designed for a long-horizon task of "Make a vanilla Americano and a slice of toast". As shown in Figure~\ref{app:fig:ben:realworld-rollout-1}, the workspace is divided into a Kitchen section and a Café section: the kitchen side contains bread slices, a spatula, a frying pan, and cookies, while the café side contains a cup, vanilla syrup bottle, classic syrup bottle, coffee machine, and water dispenser. Each section is associated with a distinct LoRA adapter, reflecting the different procedural knowledge required for kitchen-side food preparation and café-side beverage preparation. Based on this setting, the agent jointly manages heterogeneous memory over a shared task context. This environment therefore evaluates whether the agent can coordinate section-specific LoRA procedures and heterogeneous memory contents across two functionally different regions while executing a consistent long-horizon policy for preparing both the toast and the vanilla Americano.

\subsubsection{Single-Section Beverage Station}

\begin{figure}[!h]
    \centering
    \includegraphics[width=0.8\linewidth]{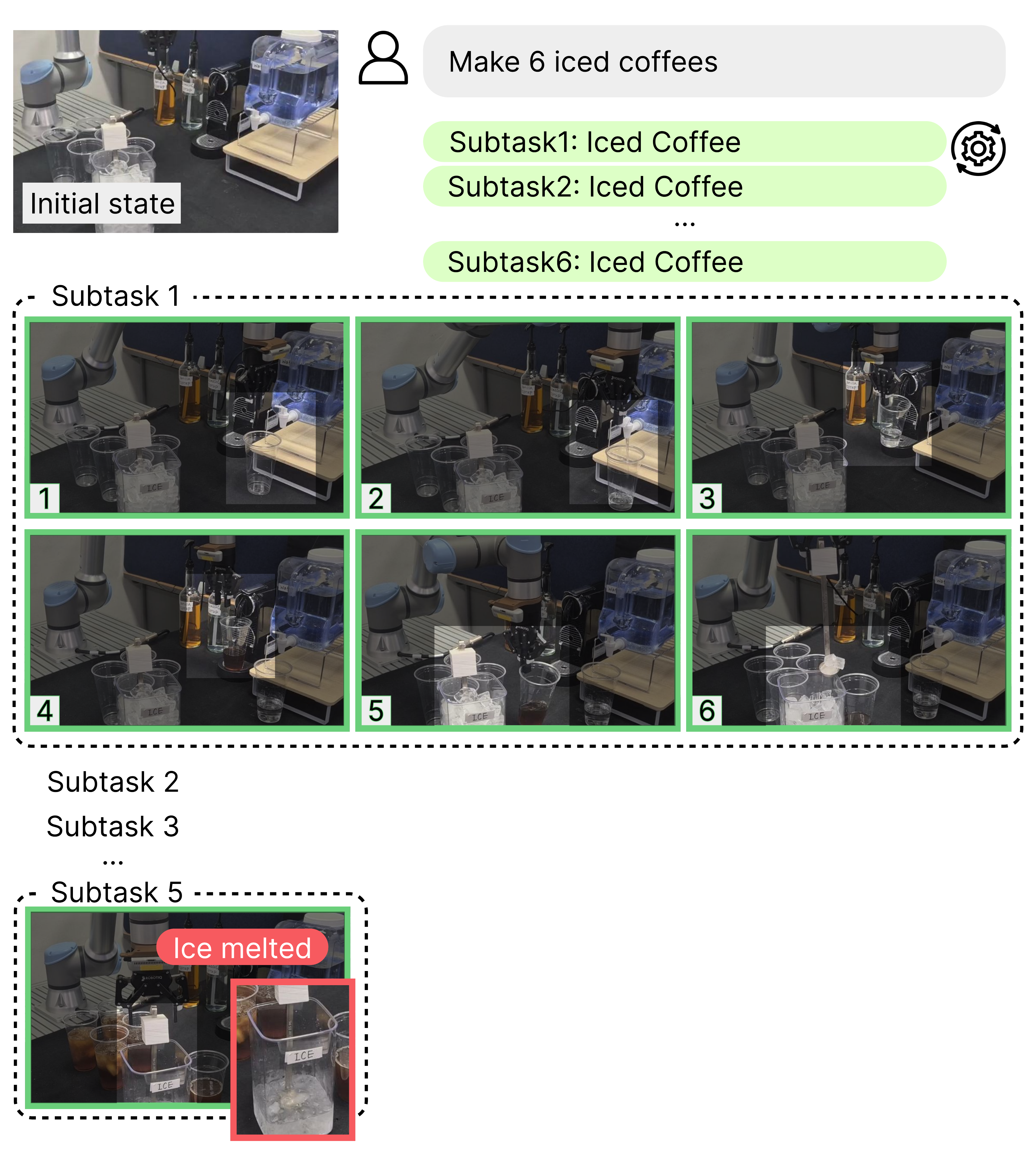}
    \caption{Real-world \textit{Single-Section Beverage Station}. The sequence is divided into six repeated coffee-making subtasks, requiring the robot to monitor the state changes of ice while preparing each beverage within a single beverage station.}
    \label{app:fig:ben:realworld-rollout-2}
\end{figure}

The second real-world environment is a \textit{single-section beverage station} designed to evaluate whether the agent can consistently update dynamic object states while performing repeated beverage-making tasks. As shown in Figure~\ref{app:fig:ben:realworld-rollout-2}, the station contains an ice container, ice scoop, syrup bottles, a coffee machine, water dispenser, and multiple cups. The task requires the agent to prepare six iced coffees sequentially within the same workspace, reusing the same beverage-making procedure across subtasks. The agent must repeatedly execute the same beverage-making procedure while monitoring changes in the shared resources, especially the ice container. Since the ice is reused across multiple subtasks and can change in quantity or condition over time, the agent needs to recognize whether sufficient ice remains, detect melting or depletion, and adapt subsequent actions accordingly.

\subsection{Baseline}
To provide a comprehensive comparison, we group the baselines by their assumptions about memory-content updates and reusable KV states during embodied inference.
\begin{itemize}[leftmargin=*, itemsep=0pt, topsep=0pt]
    \item General embodied reasoning baselines generate actions from observations, goals, and interaction history without explicit memory reconciliation.
    \begin{itemize}[leftmargin=*, itemsep=0pt, topsep=0pt]
        \item ReAct~\cite{base:react} interleaves reasoning traces and actions for step-by-step embodied decision making.
        \item Reflexion~\cite{embodied:reflexion} uses verbal self-reflection from previous failures to guide future action decisions.
    \end{itemize}

    \item Memory-augmented embodied baselines retrieve past experience or task-relevant memory to support long-horizon action selection.
    \begin{itemize}[leftmargin=*, itemsep=0pt, topsep=0pt]
        \item MemRL~\cite{mem:memrl} retrieves episodic memory from trajectories to improve embodied decision making.
    \end{itemize}

    \item KV cache reuse baselines reduce inference cost by reusing cached states under repeated or partially shared contexts.
    \begin{itemize}[leftmargin=*, itemsep=0pt, topsep=0pt]
        \item CacheBlend~\cite{serve:cacheblend} reuses non-prefix segments and selectively recomputes boundary regions.
        \item KEEP~\cite{embodied:keep} reuses KV states across embodied reasoning by invalidating affected cache portions.
    \end{itemize}

    \item Adapter-based baselines improve efficiency under procedural skill switching.
    \begin{itemize}[leftmargin=*, itemsep=0pt, topsep=0pt]
        \item LRAgent~\cite{silo:lragent} represents procedural skills with LoRA adapters and supports adapter switching during embodied inference.
    \end{itemize}
\end{itemize}
We compare $\ourmodel$ against six baselines that span both agent-level memory mechanisms and cache-level reuse strategies. All baselines share the same backbone model and decoding configuration as $\ourmodel$ (Llama-3.2-3B-Instruct on ALFWorld; Qwen3-VL-4B-Instruct on Habitat; greedy decoding) so that any difference reflects memory or cache handling rather than model capacity.

\paragraph{ReAct~\cite{base:react}.} Reasoning-and-acting prompting baseline. The agent emits a \texttt{Thought} followed by an \texttt{Action} at every step, with the conversation history accumulated as a chat trace. ReAct has no explicit memory module beyond the natural in-context history, so the cache contains only the running conversation and is rebuilt as the prompt grows.

\paragraph{Reflexion~\cite{embodied:reflexion}.} ReAct extended with verbal self-reflection on episode failure. After a failed episode, the agent generates a free-form reflection that is stored and retrieved in subsequent episodes. Reflexion tests whether self-reflective episodic feedback suffices in the absence of structured memory composition.

\paragraph{MemRL~\cite{mem:memrl}.} Memory-augmented decision making with episodic memory and Q-learned utility, but no structured KV reuse. We use the published MemRL retrieval policy (top-$k$ word-overlap retrieval with $\epsilon$-greedy exploitation) and inject retrieved trajectories as in-context successful and failed memory snippets. MemRL serves as the strongest agent-level memory baseline.

\paragraph{CacheBlend~\cite{serve:cacheblend}.} Segment-level KV reuse that fuses precomputed chunks at chunk-boundary positions with selective recompute of the boundary tokens. We adapt CacheBlend to embodied trajectories by treating each per-step memory composition as a chunked input; off-boundary updates are ignored, and at chunk-boundary steps the trailing chunks are re-prefilled with selective calibration. We use the boundary period $K = 5$ throughout, matching the recompute budget of the original system.

\paragraph{KEEP~\cite{embodied:keep}.} Embodied-memory cache management with internal-trigger invalidation. KEEP evicts memory regions that conflict with the agent's most recent action and keeps unaffected regions. We implement KEEP-style invalidation as the eviction-only subset of $\ourmodel$'s policy: the dispatcher honors only \texttt{evict} and \texttt{adapter\_switch} edits, which mirrors KEEP's self-induced invalidation scope without re-injecting the inserted or reordered content.

\paragraph{LRAgent~\cite{silo:lragent}.} Procedural memory through LoRA adapter switching with shared-attention cross-adapter cache. LRAgent reuses the prefill cache across adapter switches but does not edit the cache for declarative or working-memory updates. 
We use the same adapter pool as $\ourmodel$ while retaining LRAgent's adapter-level cache-reuse policy.

\paragraph{Faithfulness statement.}
CacheBlend, KEEP, and LRAgent are implemented according to their public descriptions and adapted to the embodied-agent evaluation setting.
Because the original systems target different serving stacks or task interfaces, we keep their core reuse assumptions unchanged while integrating them into our shared observation, action, and memory-context pipeline.
Within this harness, all baselines and $\ourmodel$ use the same observation pipeline, action filter, prompt template, backbone model, decoding setting, and admissible-command set when applicable.
Thus, performance differences primarily reflect how each method handles memory use and KV reuse under dynamically changing embodied contexts.

\subsection{Metrics}
\label{app:metrics}

We report six metrics that characterize task performance, action behavior, KV reuse, and inference efficiency.

\paragraph{Success Rate (SR).}
For ALFWorld and Habitat, SR measures binary end-to-end task completion:
\[
\mathrm{SR}
=
\frac{1}{N}
\sum_{i=1}^{N}
\mathbf{1}[\mathrm{goal\_satisfied}_i],
\]
where $N$ is the number of evaluation episodes and $\mathrm{goal\_satisfied}_i$ is the binary task-completion indicator returned by the environment.
For real-world tasks, we use goal-conditioned success rate, which measures the proportion of satisfied goal conditions within each trial, averaged across physical trials.

\paragraph{Latency.}
Latency is the average wall-clock inference time per decision step.
It is measured after the step input is finalized and includes KV reconciliation, prefill, local recomputation, and decoding.
Latency is reported in seconds per step.

\paragraph{Time-to-first-token (TTFT).}
We report TTFT as the sum of measured KV-cache update execution and fresh-tail prefill times through the first generated token.
It includes required prefill, local recomputation, and cache assembly.
For cache-reuse methods, KV ranges cached at previous steps are available before timing begins.
It excludes environment execution, memory retrieval, prompt construction, and schema-diff planning.

\paragraph{Reuse ratio.}
Reuse ratio measures the fraction of previous-step cached tokens whose KV states are preserved at the current step.
Let $S_t$ be the number of tokens in the previous cache and $C_t$ the number of those tokens whose KV states are not preserved, including evicted and locally recomputed tokens.
We average over $T$ steps with an available previous cache:
\begin{equation*}
\mathrm{Reuse}
=
1 -
\frac{1}{T}
\sum_{t=1}^{T}
\frac{C_t}{S_t}.
\end{equation*}
Inserted tokens are excluded, while RoPE-remapped KV states count as preserved.
A value of 1 indicates that all previous-step cached KV states are preserved, while 0 indicates that none are preserved.

\paragraph{Active Action Ratio.}
Active action ratio measures the fraction of generated actions that are expected to advance the task or change the environment state.
We map benchmark-specific action strings or skill API calls to a shared set of task-progressing action categories.
For episode $i$ with action sequence $\mathcal{A}_i=(a_{i,1},\ldots,a_{i,T_i})$, we compute
\begin{equation*}
\mathrm{ActiveAct}
=
\frac{1}{N}
\sum_{i=1}^{N}
\frac{1}{T_i}
\sum_{t=1}^{T_i}
\boldsymbol{1}\!\left[
a_{i,t} \in \mathcal{A}^{\mathrm{act}}
\right].
\end{equation*}
A higher value indicates that the agent more frequently performs task-progressing actions rather than passive, invalid, or redundant steps.

\paragraph{Action Diversity Ratio.}
Action diversity measures the average per-episode ratio of unique actions among all generated actions.
For episode $i$ with action sequence $\mathcal{A}_i=(a_{i,1},\ldots,a_{i,T_i})$, we compute
\begin{equation*}
\mathrm{Diversity}
=
\frac{1}{N}
\sum_{i=1}^{N}
\frac{
\left|\mathrm{Unique}(\mathcal{A}_i)\right|
}{
T_i
}.
\end{equation*}
A higher value indicates that the agent uses a broader range of actions during the trajectory.

\section{$\ourmodel$: Heterogeneous Memory Reconciliation}
\subsection{Algorithm Details}
\label{app:algorithm}

\paragraph{Overview.}
Algorithm~\ref{alg:hemer} summarizes the inference procedure of $\ourmodel$.
At each decision step, the agent first obtains memory contents from declarative, procedural, and working memory modules and forms the memory context $X_t$.
Instead of rebuilding the full KV cache for $X_t$, $\ourmodel$ compares the current memory context with the previous one using schema fields attached to each memory chunk.
The resulting reconciliation plan identifies inserted, evicted, and reordered chunks, maps these operations to KV ranges, and determines which cached states can be reused.
A cached KV range is reused only when the corresponding memory content remains compatible under the schema.
If the chunk is unchanged but only its position changes, $\ourmodel$ applies RoPE-based remapping to correct the positional phase of cached key states.
If the memory content, dependency annotation, or adapter identifier changes, the affected KV range is locally recomputed to avoid stale reuse.

\begin{algorithm}[!h]
\caption{Heterogeneous Memory Reconciliation in $\ourmodel$}
\label{alg:hemer}
\begin{algorithmic}[1]
\Require $o_t, g, h_t, X_{t-1}, \mathcal{K}_{t-1}$;
memory modules $\mathcal{M}^{\mathrm{dec}},
\mathcal{M}^{\mathrm{proc}}, \mathcal{M}^{\mathrm{work}}$
\Ensure $a_t, X_t, \mathcal{K}_t$

\State Retrieve $D_t, P_t, W_t$ using $(o_t,g,h_t)$
and construct $X_t=\Gamma(D_t,P_t,W_t)$.
\State Index chunks $\{c_i^t\}_{i=1}^{n_t}$ with persistent IDs,
positions, token spans, KV ranges, and the remaining schema metadata.

\If{$t=1$ or no previous cache is available}
    \State Prefill $X_t$ as a single causal sequence
    to obtain $\mathcal{K}_t$.
\Else
    \State Compute $\Delta_t=\mathrm{Diff}_{S}(X_{t-1},X_t)$
    by ID matching and schema checks, assigning
    \textsc{Insert}, \textsc{Evict}, \textsc{Recompute},
    \textsc{Reorder}, or \textsc{Reuse}.
    \State Propagate \textsc{Recompute} to matched direct and
    transitive dependents under declared dependencies,
    overriding \textsc{Reuse} or \textsc{Reorder}.
    \State Exclude \textsc{Evict} ranges from the current cache
    layout and determine positions from $X_t$.

    \For{each chunk $c_i^t$ in the order of $X_t$}
        \If{marked as \textsc{Insert} or \textsc{Recompute}}
            \State Prefill or recompute its KV range once at its
            current position using preceding reconciled KV ranges;
            allocate a new range for \textsc{Insert}.
        \ElsIf{marked as \textsc{Reorder}}
            \State Relocate its KV range; for eligible text-side
            keys shifted from $p_j$ to $p'_j$, apply
            $\tilde{k}_j =
            R_{\theta(p'_j)}R_{\theta(p_j)}^{-1}k_j$.
        \Else
            \State Preserve the corresponding KV range.
        \EndIf
    \EndFor

    \State Assemble $\mathcal{K}_t$ in the order of $X_t$.
\EndIf

\State Update the chunk-to-KV index.
\State Generate $a_t \sim F_\theta(\cdot \mid X_t,o_t,g)$
using $\mathcal{K}_t$.
\State \Return $a_t, X_t, \mathcal{K}_t$.
\end{algorithmic}
\end{algorithm}

\subsection{Implementation}
\label{app:implementation:system}

\paragraph{System Configuration.}
Table~\ref{tab:system-config} summarizes the software and hardware configuration used in our experiments.

\begin{table}[!h]
\centering
\caption{System configuration.}
\label{tab:system-config}
\renewcommand{\arraystretch}{1.2}
\begin{tabular}{ll}
\toprule
\textbf{Component} & \textbf{Specification} \\
\midrule
Python & 3.10.12 \\
CUDA / cuDNN & CUDA 12.4 / cuDNN 9.1 \\
PyTorch & 2.5.1 \\
Transformers & 4.46.3 \\
PEFT & 0.13.2 \\
GPU & 2$\times$ NVIDIA RTX A6000 (48GB) \\
Random Seeds & 0, 7, 42, 99, 123 \\
\bottomrule
\end{tabular}
\end{table}

\paragraph{Backbone Models.}

\begin{table}[!b]
\centering
\caption{Backbone and decoding configuration.}
\label{tab:backbone-config}
\renewcommand{\arraystretch}{1.2}
\begin{tabular}{lll}
\toprule
\textbf{Setting} & \textbf{ALFWorld} & \textbf{Habitat \& Real-world} \\
\midrule
Model & Llama-3.2-3B-Instruct & Qwen3-VL-4B-Instruct \\
Parameters & 3.2B & 4B \\
Precision & bfloat16 & bfloat16 \\
Temperature & 0.0 (greedy) & 0.0 (greedy) \\
Max new tokens & 256 & 256 \\
Observation & Text & RGB image \\
\bottomrule
\end{tabular}
\end{table}

Table~\ref{tab:backbone-config} summarizes the backbone and decoding configuration.
For multimodal backbones, RoPE-based remapping is restricted to text-side memory chunks in the memory context.
Visual tokens from the current observation are encoded normally at each decision step and are not position-remapped.

\paragraph{System architecture.}
$\ourmodel$ is implemented as an inference framework placed between the memory-augmented embodied agent and the backbone foundation model.
At each decision step, the agent retrieves declarative, procedural, and working-memory contents, forms the memory context $X_t$, and passes it to $\ourmodel$ for KV reconciliation.
The framework maintains a chunk-level memory schema, maps each memory chunk to its KV range, and reconciles the previous KV cache with the current memory context before action generation.
Environment execution and memory retrieval are handled by the agent loop, while $\ourmodel$ handles cache reuse, affected-range detection, RoPE-based remapping, local recomputation, and adapter compatibility checks.

\paragraph{Memory schema and reconciliation rules.}
Each memory content is represented as a memory chunk and stored with schema fields used for reconciliation.
The schema records each chunk's persistent ID, memory module type, semantic role, dependency annotation, temporal origin, position, token span, KV range, supporting evidence, normalized content, and adapter identifier when applicable.
$\ourmodel$ computes $\Delta_t=\mathrm{Diff}_{\mathcal{S}}(X_{t-1},X_t)$ by first matching chunks by persistent IDs and then comparing their schema fields rather than raw token overlap.
Inserted chunks are prefilling targets, evicted chunks are removed, and reordered chunks are reused only when schema fields, normalized content, and adapter identifiers remain compatible.
For schema-compatible position shifts, RoPE remapping corrects the cached key's positional phase; unsafe cases are handled by local recomputation.

\paragraph{Adapter compatibility.}
Procedural memory contents are associated with LoRA adapter configurations.
$\ourmodel$ records adapter identifiers in the memory schema and uses them as a compatibility condition for KV reuse.
If a procedural change affects an adapter-dependent range, that range is locally recomputed to avoid reusing KV states from an incompatible adapter configuration.

\paragraph{Fair comparison protocol.}
All methods use the same backbone model, task split, decoding configuration, action interface, and maximum step budget within each environment.
Memory-augmented baselines are given access to the same memory-content pool when applicable.
Cache-level baselines are evaluated on the same sequence of memory contexts and memory-content updates as $\ourmodel$.
Thus, the comparison isolates how each method processes the memory context and reuses or recomputes KV states.

\subsection{Prompt Templates}
\label{app:prompts}

We document the prompt templates used in our experiments.
The system prompt, task brief, action grammar, and observation-action format are shared across methods within each benchmark.
Methods differ only in whether and how they use the provided memory contents, and in how their serving policy reuses or recomputes KV states.
This ensures that performance differences are not caused by benchmark-specific prompt engineering.
Although the prompt template is shown in full, cache-reuse methods do not prefill the entire rendered prompt at every decision step.
TTFT includes KV cache updates, required prefill, and local recomputation up to the first generated token, excluding memory retrieval, prompt construction, and schema-diff planning.

\paragraph{Shared  System Prompt.}
\label{app:prompts:sys}

The same system prompt is used by ReAct, Reflexion, MemRL, CacheBlend, KEEP, LRAgent, and \ourmodel{}.

\begin{quote}\small\ttfamily
Interact with a household to solve a task.
Imagine you are an intelligent agent in a household environment and your target is to perform actions to complete the task goal.
At the beginning of your interactions, you will be given the detailed description of the current environment and your goal to accomplish.\\[2pt]
For each turn, you will be given the observation from the last turn.
You should first think about the current condition and plan your future actions, and then output your action for this turn.
Your output must strictly follow this format:\\
``Thought: your thoughts.\textbackslash nAction: your next action''.\\[2pt]
The available actions are:\\
1. go to \{recep\}\\
2. take \{obj\} from \{recep\}\\
3. move \{obj\} to \{recep\}\\
4. open \{recep\}\\
5. close \{recep\}\\
6. use \{obj\}\\
7. clean \{obj\} with \{recep\}\\
8. heat \{obj\} with \{recep\}\\
9. cool \{obj\} with \{recep\}\\[2pt]
where \{obj\} and \{recep\} correspond to objects and receptacles.
After each turn, the environment will provide feedback.
If the environment outputs ``Nothing happened'', the previous action is invalid and you should try another valid action.\\[2pt]
Your response should use the following format:\\
Thought: \textless your thoughts\textgreater\\
Action: \textless your next action\textgreater
\end{quote}

The same Thought-Action output format is used for all methods, including cache-reuse baselines, so that differences are not caused by response-format prompting.

\paragraph{Memory Context Block.}
\label{app:prompts:mem}

For memory-augmented methods, retrieved memory contents are rendered into a typed memory context block.
For \ourmodel{}, this block corresponds to the memory context $X_t$ used by the reconciliation policy.
The same text-side memory context format is used across benchmarks, while visual observations are provided through the multimodal input interface when applicable.

\begin{quote}\small\ttfamily
{[}Memory context $X_t${]}\\
{[}Declarative memory{]}\\
\ \ - \textless declarative memory content 1\textgreater\\
\ \ - \textless declarative memory content 2\textgreater\\
\ \ \ldots\\
{[}Procedural memory{]}\\
\ \ - \textless procedural memory content\textgreater\\
{[}Working memory{]}\\
\ \ - \textless working-memory summary 1\textgreater\\
\ \ - \textless working-memory summary 2\textgreater\\
\ \ \ldots
\end{quote}

At each decision step, the block is rendered from the current memory contents and their schema fields.
Memory-content updates, such as insertion, eviction, and reordering, appear to the foundation model through the updated memory context, while procedural adapter changes are tracked as schema metadata for KV compatibility.
For cache-level baselines, we use the same rendered memory context when applicable, but each method applies its own cache reuse policy.
For \ourmodel{}, the rendered chunks are additionally aligned with KV ranges through the memory schema, enabling schema-guided KV reconciliation.
Adapter identifiers are stored as schema metadata for compatibility checking and are not used as additional natural-language hints unless required by a baseline interface.

\paragraph{Task Brief.}
\label{app:prompts:task}

The first user turn specifies the task type, goal, and initial admissible commands.
The format is shared across methods:

\begin{quote}\small\ttfamily
You are solving a \{task\_type\} task.\\
Goal: \{task\}\\
Initial admissible commands: \{cmd\_1\}, \{cmd\_2\}, \ldots, \{cmd\_K\}
\end{quote}

We use the same admissible-command formatting across methods.
When the admissible-command list is long, it is truncated using the same limit for all methods.
Subsequent observation turns include the dynamically updated admissible-command list.

\paragraph{Observation-Action History.}
\label{app:prompts:hist}

After the task brief, each step appends a user turn of the form:

\begin{quote}\small\ttfamily
Observation: \{obs\}
\end{quote}

The assistant response follows the shared ReAct-style format:

\begin{quote}\small\ttfamily
Thought: \textless reasoning about the current state\textgreater\\
Action: \textless next action\textgreater
\end{quote}

The conversation history persists within an episode so that each method has access to previous observations and actions.
Long observations are truncated using the same character budget across all methods.
The memory context block may change across steps as memory contents are updated, but past assistant turns are not rewritten.

\paragraph{Retrieved Trajectory Memory.}
\label{app:prompts:trajectory}

For methods that use episodic or trajectory memory, we optionally include a compact retrieved trajectory from the same task type.
This block is used only for methods whose baseline design includes retrieved experience, such as MemRL and memory-augmented variants of \ourmodel{}.
The retrieved trajectory format is:

\begin{quote}\small\ttfamily
--- RETRIEVED TRAJECTORY MEMORY ---\\
Task: \{retrieved task description\}\\
Action sequence: \{action\_1\} \textgreater\ \{action\_2\} \textgreater\ \ldots
\end{quote}

When this block is used, the same retrieval source and formatting are applied to all methods that support trajectory memory.
This isolates the effect of memory reconciliation and KV reuse from differences in retrieved-memory quality.

\paragraph{Fairness note.}
Prompt templates are shared at the level of system instruction, action grammar, task brief, and observation-action history.
When a method supports memory input, it receives the same rendered memory-content format; methods that do not use memory ignore this block by design.
For cache-reuse comparisons, the same memory context sequence is used whenever the method can consume it, and differences arise from how KV states are reused or recomputed.

\subsection{Hyperparameters}
\label{app:hyperparameters}

\begin{table}[!h]
\centering
\caption{Hyperparameters used by $\ourmodel$ across experiments.}
\label{tab:hyperparameters}
\renewcommand{\arraystretch}{1.15}
\begin{tabular}{ll}
\toprule
\textbf{Hyperparameter} & \textbf{Value} \\
\midrule
Decoding temperature & $0$ \\
Max new tokens & $256$ \\
Maximum ALFWorld steps & $30$ \\
Maximum Habitat steps & $50$ \\
LoRA rank & $8$ \\
\bottomrule
\end{tabular}
\end{table}

\paragraph{Reproducibility notes.}
Local recomputation is included in the reported latency and TTFT.
Active action ratio and action diversity are computed from generated action sequences, as described in Appendix~\ref{app:metrics}.

\section{Additional experimental result}

\subsection{Main table}
Tables~\ref{tab:app:main:alf}, \ref{tab:app:main:habitat}, and~\ref{tab:app:main:real-world} provide the full per-environment results used in the main evaluation.
We provide additional analysis for each environment below.
\paragraph{ALFWorld results.}
Table~\ref{tab:app:main:alf} reports the full ALFWorld results.
$\ourmodel$ achieves the highest SR at $71.43\%$, outperforming the strongest baseline MemRL by $18.57$ percentage points.
It also reduces latency from $0.90$ seconds per step for MemRL to $0.18$ seconds per step.
This indicates that the improvement is not only due to better memory use, but also to lower inference cost under changing memory contexts.
ReAct and Reflexion perform worse because they rely mainly on prompting or self-reflection without explicit structured reuse of memory contents.
MemRL improves over these prompting baselines by retrieving episodic information from previous trajectories, but it still processes the retrieved memory context as a conventional input.
As a result, it does not reduce the cost of repeatedly processing updated memory contents.

The cache-reuse baselines reduce latency compared with memory-only baselines, but their task success remains lower than $\ourmodel$.
CacheBlend and KEEP achieve lower latency than MemRL, but their reuse ratios remain around $50\%$, substantially below the $75.30\%$ reuse achieved by $\ourmodel$.
LRAgent also reuses part of the computation through procedural adapter switching, but its lower SR and diversity suggest that procedural reuse alone is insufficient when declarative and working-memory contents change across steps.
By contrast, $\ourmodel$ achieves both the highest reuse ratio and the highest action behavior metrics.
This shows that schema-indexed memory chunks help preserve compatible KV states while still allowing the agent to use a broader and more task-relevant set of actions.
Overall, the ALFWorld results demonstrate that $\ourmodel$ is especially effective in text-only environments where admissible actions, object locations, and working-memory summaries change frequently across decision steps.

\begin{table*}[h]                                                                                                  
    \centering                                              
    \caption{Main results on ALFWorld.}       
    \label{tab:app:main:alf}
    \begin{adjustbox}{width=.99\textwidth}
    \begin{tabular}{lcccccc}                                                                                         
    \toprule                                                                                                         
    \textbf{Method} & \textbf{SR (\%)} & \textbf{Latency (s/step)} & \textbf{Reuse ratio(\%)} &  \textbf{Active ratio (\%)} & \textbf{Diversity ratio(\%)} & \textbf{TTFT (ms)} \\                                                                            
    \midrule                              
    $\ourmodel$& $71.43 \pm 2.05$ & $0.18 \pm 0.02$  & $75.30 \pm 4.33$ & $67.69 \pm 2.27$ & $71.43 \pm 3.32$ & $21.59 \pm 2.53$ \\
    ReAct      & $47.14 \pm 6.39$ & $0.88 \pm 0.04$  & --               & $52.04 \pm 2.45$ & $64.95 \pm 8.94$ & $23.32 \pm 3.24$ \\
    Reflexion  & $44.29 \pm 5.74$ & $0.98 \pm 0.05$  & --               & $45.57 \pm 4.30$ & $64.38 \pm 6.12$ & $23.12 \pm 2.89$ \\
    MemRL      & $52.86 \pm 4.83$ & $0.90 \pm 0.04$  & --               & $46.26 \pm 5.16$ & $65.05 \pm 4.48$ & $22.88 \pm 2.71$ \\
    CacheBlend & $48.39 \pm 4.29$ & $0.48 \pm 0.04$  & $50.30 \pm 2.33$ & $46.01 \pm 3.67$ & $58.49 \pm 2.92$ & $22.35 \pm 3.22$ \\
    KEEP       & $51.81 \pm 3.92$ & $0.46 \pm 0.06$  & $48.30 \pm 2.33$ & $50.40 \pm 4.92$ & $65.01 \pm 4.95$ & $22.39 \pm 3.24$ \\
    LRAgent    & $46.43 \pm 3.19$ & $0.52 \pm 0.03$  & $48.50 \pm 3.33$ & $51.17 \pm 2.31$ & $49.98 \pm 3.06$ & $22.38 \pm 1.99$ \\
    \bottomrule                                                                                                      
    \end{tabular}%
    \end{adjustbox}
\end{table*}   

\paragraph{Habitat results.}
Table~\ref{tab:app:main:habitat} reports the full results on EmbodiedBench Habitat.
$\ourmodel$ achieves $76.27\%$ SR, improving over MemRL by $14.47$ percentage points and over the best cache baseline KEEP by $16.87$ percentage points.
It also obtains the lowest latency at $1.82$ seconds per step, compared with $2.68$ seconds for MemRL and $1.97$--$2.09$ seconds for the cache-reuse baselines.
This is important because Habitat introduces RGB observations and high-level skill transitions, making the memory context depend not only on text history but also on visual grounding and changing object states.
The result therefore suggests that $\ourmodel$ is not limited to text-only memory updates, but can also support multimodal embodied inference where visual observations update the relevant memory contents.

The behavior metrics further show that $\ourmodel$ maintains stronger action-level performance.
It achieves the highest active action ratio at $71.38\%$ and the highest action diversity at $56.29\%$.
MemRL obtains relatively high diversity, but its latency remains high because it does not reuse KV states.
CacheBlend, KEEP, and LRAgent reduce inference time relative to MemRL, but their lower SR and reuse ratios show that their reuse mechanisms are less robust to multimodal memory-content changes.
CacheBlend relies on reusable non-prefix segments, KEEP invalidates affected regions, and LRAgent focuses on procedural adapter switching.
These mechanisms do not fully address the case where visual facts, working-memory summaries, and procedural skill choices change together.
By contrast, $\ourmodel$ maintains a higher reuse ratio by matching schema-indexed chunks and locally updating affected KV ranges.
This explains why it improves both task success and inference efficiency on Habitat.

\begin{table*}[h]
    \centering
    \caption{Main results on Habitat.}
    \label{tab:app:main:habitat}
    \begin{adjustbox}{width=.99\textwidth}
    \begin{tabular}{lcccccc}                                                                                         
    \toprule
    \textbf{Method} & \textbf{SR (\%)} & \textbf{Latency (s/step)} & \textbf{Reuse ratio(\%)} &  \textbf{Active ratio (\%)} & \textbf{Diversity ratio(\%)} & \textbf{TTFT (ms)} \\                                                                            
    \midrule
    $\ourmodel$& $76.27 \pm 3.32$ & $1.82 \pm 0.03$ & $72.81 \pm 4.24$ & $71.38 \pm 3.29$ & $56.29 \pm 2.82$ & $38.29 \pm 3.29$ \\
    ReAct      & $52.00 \pm 3.32$ & $2.41 \pm 0.07$ & --               & $51.81 \pm 6.33$ & $42.03 \pm 2.12$ & $44.30 \pm 5.78$ \\
    Reflexion  & $50.47 \pm 5.40$ & $2.38 \pm 0.18$ & --               & $49.66 \pm 8.71$ & $39.27 \pm 3.04$ & $44.34 \pm 4.90$ \\
    MemRL      & $61.80 \pm 2.92$ & $2.68 \pm 0.12$ & --               & $57.55 \pm 4.78$ & $54.36 \pm 1.31$ & $43.89 \pm 3.29$ \\
    CacheBlend & $56.33 \pm 2.24$ & $1.97 \pm 0.03$ & $46.25 \pm 3.29$ & $51.39 \pm 3.29$ & $53.20 \pm 2.30$ & $39.82 \pm 4.83$ \\
    KEEP       & $59.40 \pm 2.24$ & $2.09 \pm 0.02$ & $43.05 \pm 4.12$ & $64.58 \pm 1.75$ & $52.40 \pm 1.20$ & $40.71 \pm 6.08$ \\
    LRAgent    & $57.47 \pm 3.54$ & $2.03 \pm 0.31$ & $42.96 \pm 3.98$ & $58.34 \pm 4.53$ & $44.28 \pm 3.39$ & $40.57 \pm 5.06$ \\
    \bottomrule
    \end{tabular}%
    \end{adjustbox}
\end{table*}

\paragraph{RealWorld-Tabletop results.}
Table~\ref{tab:app:main:real-world} reports results on the real-world tabletop manipulation setting.
$\ourmodel$ achieves the highest SR at $83.20\%$, outperforming MemRL by $18.92$ percentage points.
It also reduces latency from $2.94$ to $1.87$ seconds per step, showing that the efficiency benefit remains under real sensor observations and physical manipulation.
Unlike simulation, the real-world setting introduces perception noise, object pose changes, execution uncertainty, and task-state changes caused by robot actions.
These factors can change which memory contents are relevant at each decision step.
The strong result of $\ourmodel$ indicates that memory-aware KV reconciliation remains useful when memory-content updates arise from physical interaction rather than only from simulated state transitions.

The baseline trends show a similar pattern to the simulation results.
MemRL improves task success over some cache-only baselines by using retrieved memory, but it has the highest latency because it lacks structured KV reuse.
CacheBlend and LRAgent reduce latency relative to MemRL, but their lower SR suggests that preserving some cached computation is not enough when the action context changes after perception and manipulation updates.
KEEP achieves a stronger SR than CacheBlend and LRAgent, but it has lower reuse and higher latency than $\ourmodel$, reflecting the cost of broader invalidation.
By contrast, $\ourmodel$ achieves the highest reuse ratio at $72.18\%$, the highest active action ratio at $75.45\%$, and the highest action diversity at $64.85\%$.
This indicates that $\ourmodel$ preserves compatible KV states while continuing to generate varied and task-progressing actions.
Overall, the real-world results support the claim that heterogeneous-memory KV reconciliation improves both robustness and efficiency in embodied tasks with dynamic observations and physical state changes.

\begin{table*}[h]
    \centering
    \caption{Main results on Real-world.}
    \label{tab:app:main:real-world}
    \begin{adjustbox}{width=.99\textwidth}
    \begin{tabular}{lcccccc}                                                                                         
    \toprule
    \textbf{Method} & \textbf{SR (\%)} & \textbf{Latency (s/step)} & \textbf{Reuse ratio(\%)} &  \textbf{Active ratio (\%)} & \textbf{Diversity ratio(\%)} & \textbf{TTFT (ms)} \\                                                                            
    \midrule
    $\ourmodel$   & $83.20 \pm 4.39$ & $1.87 \pm 0.12$ & $72.18 \pm 3.28$ & $75.45 \pm 3.92$ & $64.85 \pm 2.24$ & $57.29 \pm 5.39$ \\
    MemRL      & $64.28 \pm 5.25$ & $2.94 \pm 0.32$ & --               & $61.04 \pm 4.74$ & $58.19 \pm 2.73$ & $83.20 \pm 7.18$ \\
    CacheBlend & $51.21 \pm 4.02$ & $2.14 \pm 0.12$ & $54.02 \pm 4.30$ & $54.27 \pm 3.82$ & $54.37 \pm 4.19$ & $59.29 \pm 4.87$ \\
    KEEP       & $63.40 \pm 3.27$ & $2.43 \pm 0.24$ & $48.64 \pm 4.24$ & $62.03 \pm 3.68$ & $56.20 \pm 3.28$ & $63.93 \pm 5.92$ \\
    LRAgent    & $53.43 \pm 4.18$ & $2.21 \pm 0.18$ & $50.22 \pm 3.82$ & $56.30 \pm 2.39$ & $49.97 \pm 1.29$ & $63.21 \pm 7.38$ \\
    \bottomrule
    \end{tabular}%
    \end{adjustbox}
\end{table*}

\subsection{Analysis}
\paragraph{Cost-analysis raw data.}
The TTFT values are measured after the step input is finalized and reusable KV ranges are available, so they reflect the incremental cache-processing and first-token cost rather than full prompt construction or environment execution time.
Tables~\ref{tab:cost:ttft_memlen} and~\ref{tab:cost:reuse_rho} provide the raw numerical values used for the cost-efficiency analysis in Figure~\ref{fig:eval:ana:comb}.
Table~\ref{tab:cost:ttft_memlen} reports TTFT as the memory length increases from 5 to 50 chunks.
Full Rebuild shows a nearly monotonic increase in TTFT, reaching $148.59$ ms at 50 chunks, because the entire memory context must be prefilled at each decision step.
CacheBlend, KEEP, and LRAgent reduce this cost by reusing part of the cached states, but their TTFT still grows substantially as the memory context becomes longer.
In contrast, $\ourmodel$ maintains the lowest TTFT across all memory lengths, reaching $73.49$ ms at 50 chunks.
This indicates that chunk-level reconciliation reduces the amount of memory context that must be reprocessed as the memory length grows.

Table~\ref{tab:cost:reuse_rho} reports KV reuse ratio as the invalidated memory ratio $\rho$ increases.
Full Rebuild has zero reuse by construction.
CacheBlend, KEEP, and LRAgent preserve some cached states when $\rho$ is small, but their reuse ratios decrease rapidly as more memory regions are invalidated.
For example, at $\rho=0.5$, their reuse ratios drop to $0.184$, $0.152$, and $0.134$, respectively.
By comparison, $\ourmodel$ maintains a reuse ratio of $0.453$ at the same invalidation level.
This shows that $\ourmodel$ can preserve compatible KV states even when a substantial fraction of the memory context is perturbed.
The gap comes from grounding cache reuse in schema-indexed memory chunks rather than relying only on prefix stability, segment boundaries, or adapter-level reuse.

Together, these results show that $\ourmodel$ improves inference efficiency from two complementary directions.
First, it lowers TTFT under long memory contexts by avoiding full cache reconstruction.
Second, it maintains higher KV reuse under memory perturbation by locally updating only affected KV ranges.
These trends support the main claim that memory-aware KV reconciliation can reduce redundant inference computation while preserving reusable memory contents across decision steps.

\begin{table*}[h]
\centering
\caption{Analysis of TTFT (ms) under increasing memory length. Column headings indicate the number of memory chunks. Values are mean $\pm$ std.}
\label{tab:cost:ttft_memlen}
\resizebox{\textwidth}{!}{%
\begin{tabular}{lcccccc}
\toprule
\textbf{Method} & \textbf{5} & \textbf{10} & \textbf{20} & \textbf{30} & \textbf{40} & \textbf{50} \\
\midrule
Full Rebuild & $22.88 \pm 1.68$ & $39.99 \pm 1.87$ & $55.54 \pm 4.30$ & $90.20 \pm 4.39$ & $112.46 \pm 7.27$ & $148.59 \pm 8.20$ \\
CacheBlend & $22.80 \pm 3.23$ & $23.92 \pm 1.58$ & $43.39 \pm 4.01$ & $65.93 \pm 4.29$ & $78.23 \pm 8.54$ & $92.15 \pm 8.30$ \\
KEEP & $22.35 \pm 1.99$ & $23.89 \pm 1.98$ & $43.38 \pm 8.07$ & $66.84 \pm 6.19$ & $78.35 \pm 7.42$ & $90.82 \pm 8.11$ \\
LRAgent & $22.39 \pm 3.24$ & $26.39 \pm 2.95$ & $46.30 \pm 3.27$ & $68.30 \pm 4.47$ & $82.24 \pm 5.28$ & $102.39 \pm 8.42$ \\
$\ourmodel$ & $\mathbf{21.60 \pm 2.53}$ & $\mathbf{21.91 \pm 1.83}$ & $\mathbf{27.37 \pm 5.44}$ & $\mathbf{43.49 \pm 3.29}$ & $\mathbf{58.53 \pm 4.11}$ & $\mathbf{73.49 \pm 5.34}$ \\
\bottomrule
\end{tabular}%
}
\end{table*}

\begin{table*}[h]
\centering
\caption{Analysis on KV reuse ratio under increasing invalidated memory ratio $\rho$. Values are mean $\pm$ std.}
\label{tab:cost:reuse_rho}
\resizebox{\textwidth}{!}{%
\begin{tabular}{lcccccc}
\toprule
\textbf{Method} & $\boldsymbol{\rho=0.0}$ & $\boldsymbol{\rho=0.1}$ & $\boldsymbol{\rho=0.2}$ & $\boldsymbol{\rho=0.3}$ & $\boldsymbol{\rho=0.4}$ & $\boldsymbol{\rho=0.5}$ \\
\midrule
Full Rebuild & $0.000 \pm 0.000$ & $0.000 \pm 0.000$ & $0.000 \pm 0.000$ & $0.000 \pm 0.000$ & $0.000 \pm 0.000$ & $0.000 \pm 0.000$ \\
CacheBlend & $0.503 \pm 0.023$ & $0.387 \pm 0.032$ & $0.385 \pm 0.040$ & $0.324 \pm 0.034$ & $0.313 \pm 0.083$ & $0.184 \pm 0.053$ \\
KEEP & $0.483 \pm 0.023$ & $0.427 \pm 0.032$ & $0.385 \pm 0.010$ & $0.324 \pm 0.034$ & $0.201 \pm 0.033$ & $0.152 \pm 0.094$ \\
LRAgent & $0.485 \pm 0.033$ & $0.463 \pm 0.023$ & $0.323 \pm 0.081$ & $0.292 \pm 0.081$ & $0.282 \pm 0.083$ & $0.134 \pm 0.042$ \\
$\ourmodel$ & $\mathbf{0.753 \pm 0.043}$ & $\mathbf{0.709 \pm 0.046}$ & $\mathbf{0.652 \pm 0.057}$ & $\mathbf{0.605 \pm 0.057}$ & $\mathbf{0.504 \pm 0.049}$ & $\mathbf{0.453 \pm 0.021}$ \\
\bottomrule
\end{tabular}%
}
\end{table*}

\paragraph{Hidden-State Drift Analysis.}
\label{app:drift_analysis}

\begin{figure*}[h]
\begin{center}
\includegraphics[width=0.99\linewidth]{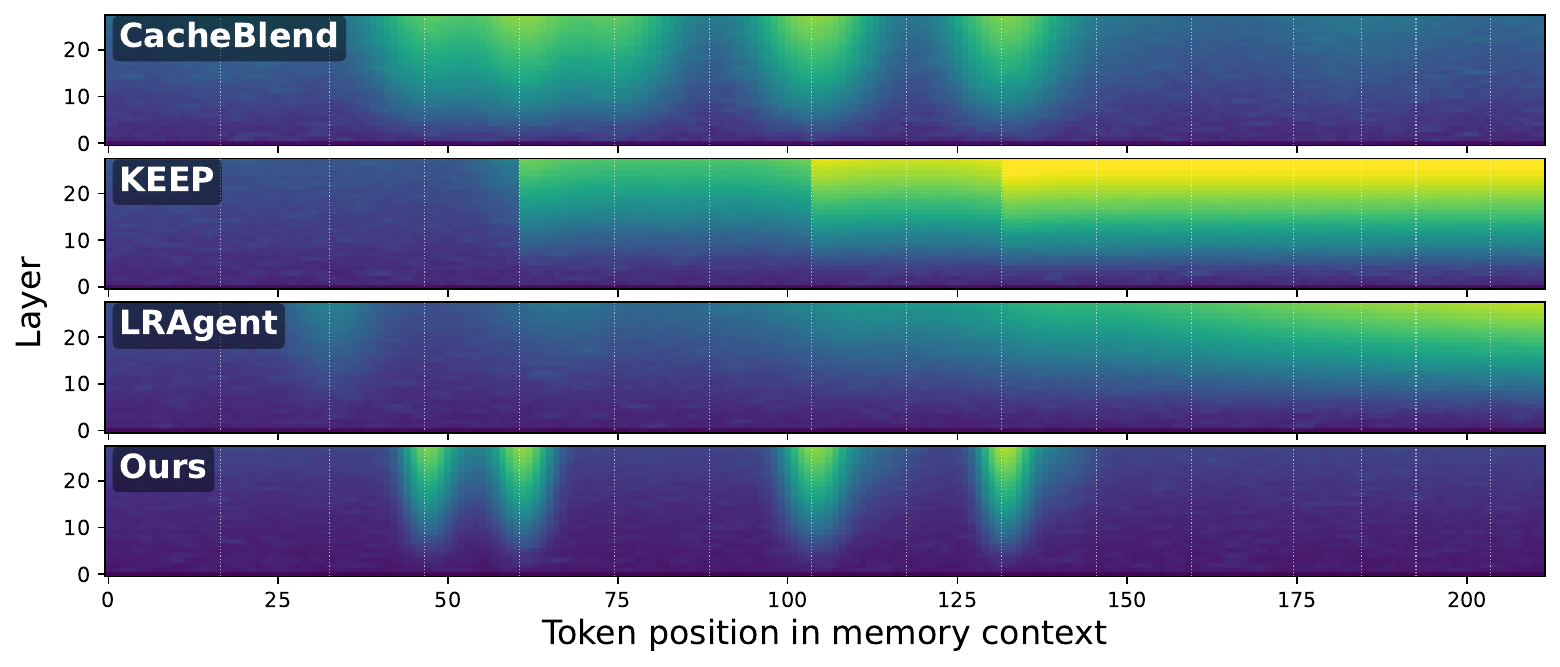}
\end{center}
\caption{
Analysis on  Hidden-state drift from Full Rebuild shows that $\ourmodel$ localizes drift around affected chunks, while baselines show broader boundary-level or downstream drift.
}
\label{fig:eval:ana:attn}
\end{figure*}

We further analyze representation-level robustness under memory-content updates.
While task success measures whether the final action trajectory satisfies the goal, it does not directly show whether KV reuse destabilizes internal model representations.
To inspect this effect, we compare each cache-reuse method against Full Rebuild, which recomputes the full KV cache after every memory-content update.
For each method, we measure hidden-state drift at each layer and token position:
\begin{equation}
\mathrm{Drift}_{\ell,i}
=
\left\|
h^{\mathrm{method}}_{\ell,i}
-
h^{\mathrm{full}}_{\ell,i}
\right\|_2,
\end{equation}
where $h^{\mathrm{method}}_{\ell,i}$ is the hidden state at layer $\ell$ and token position $i$ after applying the method's cache update rule, and $h^{\mathrm{full}}_{\ell,i}$ is the corresponding hidden state from Full Rebuild.
The heatmap in Figure~\ref{fig:eval:ana:attn} visualizes this drift across layers and token positions.
Brighter regions indicate larger representation mismatch from Full Rebuild.

The drift patterns reveal how each method propagates memory-content updates through the cached representation.
CacheBlend produces drift around reusable segment boundaries.
This suggests that segment-level reuse is sensitive to boundary mismatch: even when a non-prefix segment is reused, local changes near the segment boundary can perturb nearby reused states.
KEEP shows broader downstream propagation.
Because KEEP invalidates affected regions and preserves later cache regions under a coarser update rule, representation mismatch can propagate to later token positions after a memory-content update.
LRAgent shows gradual drift accumulation over later positions.
This pattern indicates that adapter-level reuse alone does not fully account for updates to declarative or working-memory contents, so representation mismatch accumulates when the memory context changes across decision steps.

In contrast, $\ourmodel$ confines high-drift regions to a small number of affected memory chunks.
This pattern follows from the schema-indexed reconciliation rule.
A KV range is reused only when the corresponding memory content remains compatible under the memory schema, including its normalized content, dependency annotation, and adapter identifier.
When the update is limited to a position shift, RoPE remapping corrects the positional phase of cached key states.
When the memory content, dependency annotation, or adapter identifier changes, $\ourmodel$ locally recomputes the affected KV range rather than reusing stale states.
In the evaluated settings, representation deviations remain comparatively small outside the affected memory chunks.

This analysis supports the robustness claim of $\ourmodel$.
The benefit is not only that more KV states are reused, but that reused states remain localized and compatible with the current memory context.
Thus, $\ourmodel$ reduces redundant recomputation while limiting broad hidden-state drift under the evaluated memory-content updates.

\paragraph{Deviation Accumulation over Successive Memory-Content Updates.}
\label{app:accum}
While the hidden-state drift analysis above measures deviation after individual memory-content updates, we further examine whether the deviation accumulates over successive updates.
Table~\ref{tab:app:accum} reports the hidden-state cosine similarity to Full Rebuild over $30$ successive memory-content updates whose operation distribution follows the ALFWorld episodes.
Naive reuse preserves all cached states without reconciliation, so its deviation accumulates with no mechanism to reset it, ending at $0.949$; its transient rise at step $20$ comes from evictions that remove previously changed content together with its accumulated deviation, not from any correction of cached states.
CacheBlend recomputes trailing chunks with a boundary period of $5$ updates, so its values fluctuate with the distance to the most recent recomputation.
$\ourmodel$ reaches its lowest value in the middle of the horizon and recovers to $0.974$, because every content change and adapter switch triggers recomputation that refreshes the affected KV ranges and limits accumulated deviation.

\begin{table*}[h]
\centering
\caption{Analysis on Deviation accumulation over successive memory-content updates.
Values are hidden-state cosine similarity to Full Rebuild, mean $\pm$ std over five update sequences.}
\label{tab:app:accum}
\resizebox{\textwidth}{!}{%
\begin{tabular}{lccccc}
\toprule
\textbf{Method} & \textbf{Step 1} & \textbf{Step 5} & \textbf{Step 10} & \textbf{Step 20} & \textbf{Step 30} \\
\midrule
Naive reuse & $0.993 \pm 0.007$ & $0.974 \pm 0.012$ & $0.948 \pm 0.012$ & $0.971 \pm 0.015$ & $0.949 \pm 0.016$ \\
CacheBlend & $0.993 \pm 0.007$ & $0.974 \pm 0.012$ & $0.947 \pm 0.012$ & $0.980 \pm 0.004$ & $0.972 \pm 0.009$ \\
$\ourmodel$ & $0.993 \pm 0.007$ & $0.989 \pm 0.006$ & $0.969 \pm 0.008$ & $0.979 \pm 0.003$ & $0.974 \pm 0.007$ \\
\bottomrule
\end{tabular}%
}
\end{table*}

\paragraph{Robustness to distractor action candidates.}
The results show that $\ourmodel$ remains stable under moderate distractor levels.
From $0\%$ to $100\%$ distractor noise, SR decreases only slightly, from $71.43\%$ to $68.70\%$, while latency increases from $0.18$ to $0.42$ seconds per step.
Although the reuse ratio decreases as more irrelevant candidates are injected, $\ourmodel$ still preserves more than half of the cached states at $100\%$ noise and keeps TTFT within a moderate range.
This suggests that the framework can maintain task-relevant decision behavior under noisy action contexts while limiting unnecessary cache reconstruction.
At higher distractor levels, SR drops sharply and TTFT increases, indicating that excessive irrelevant candidates eventually dominate action selection and reduce cache reuse.
\begin{table*}[h]
\centering
\caption{
Analysis on Robustness under distractor action candidates.
We inject irrelevant admissible-action candidates into the decision context and measure task success, action behavior, inference time, KV reuse, and TTFT.
$\ourmodel$ remains stable under moderate distractor ratios, while extreme distractor levels degrade action selection and reduce reusable cache states.}
\label{tab:r1_scoping_robustness}
\resizebox{\linewidth}{!}{%
\begin{tabular}{lcccccc}
\toprule
\textbf{Noise} & \textbf{SR (\%)} & \textbf{Latency (s/step)} & \textbf{Reuse ratio(\%)} & \textbf{Active ratio(\%)} & \textbf{Diversity ratio(\%)} & \textbf{TTFT (ms)}\\
\midrule
0\%& $\mathbf{71.43 \pm 2.05}$ & $\mathbf{0.18 \pm 0.02}$ & $\mathbf{75.30 \pm 4.33}$ & $\mathbf{67.69 \pm 2.27}$ & $\mathbf{71.43 \pm 3.32}$ & $\mathbf{21.59 \pm 2.53}$ \\
30\%         & 71.06 $\pm$ 2.43 & 0.21 $\pm$ 0.12 & 70.23 $\pm$ 1.31 & 76.09 $\pm$ 1.91 & 82.29 $\pm$ 2.01 & $25.94 \pm 3.24$\\
60\%         & 70.73 $\pm$ 3.89 & 0.23 $\pm$ 0.08 & 63.72 $\pm$ 2.04 & 64.60 $\pm$ 0.71 & 77.42 $\pm$ 1.21 & $30.89 \pm 4.44$\\
100\%        & 68.70 $\pm$ 2.86 & 0.42 $\pm$ 0.02 & 55.44 $\pm$ 4.12 & 45.83 $\pm$ 2.31  & 69.58 $\pm$ 1.43 & $48.02 \pm 5.10$\\
\midrule
150\%        & 53.29 $\pm$ 5.55 & 0.48 $\pm$ 0.10 & 42.42 $\pm$ 2.60 & 35.98 $\pm$ 2.39 & 53.42 $\pm$ 2.24 & $52.08 \pm 5.40$\\
200\%        & 31.01 $\pm$ 2.06 & 0.62 $\pm$ 0.24 & 31.29 $\pm$ 8.20 & 34.69 $\pm$ 2.07 & 43.92 $\pm$ 1.20 & $65.19 \pm 3.29$\\
300\%        & 15.30 $\pm$ 1.28 & 0.93 $\pm$ 0.11 & 25.64 $\pm$ 6.64 & 33.13$\pm$ 1.92 & 33.23 $\pm$ 1.42 & $83.24 \pm 5.41$\\
\bottomrule
\end{tabular}%
}
\end{table*}

\paragraph{Approximation quality of non-prefix KV reuse.}
Since reused KV states retain the influence of their previous preceding context, chunk-level KV reconciliation approximates full recomputation (Sec.~\ref{sec:method:kv}).
Table~\ref{tab:ana:approx} compares it with Full Rebuild across memory-content update patterns using five memory contexts composed as in the ALFWorld experiments.

K cos measures reused-key cosine similarity, and hidden cos measures last-layer hidden-state cosine similarity over a fixed action-decision prompt appended after the memory context.
KL measures the divergence between next-token distributions at the prompt's final position.
We report two complementary agreement metrics against Full Rebuild.
Top-1 agreement measures the fraction of evaluation cases in which the highest-probability next token is identical between the two methods.
Top-5 overlap measures the average fraction of shared tokens between their top-5 next-token sets, computed as the intersection size divided by five.
Unlike hit@5, Top-5 overlap measures agreement between candidate sets rather than the inclusion of a reference token, and can therefore be lower than Top-1 agreement.
Content changes and adapter switches trigger recomputation of affected KV ranges, yielding KL below $0.01$.
Insertion, eviction, and reordering reuse downstream ranges despite changes to their preceding context, yet hidden-state cosine remains at least $0.984$.
Eviction produces the largest output deviation, while reordering produces the largest key-state deviation.
Across $30$ successive updates following the ALFWorld operation distribution, recomputation on content changes and adapter switches limits accumulated deviation, yielding a final hidden-state cosine of $0.974$, compared with $0.949$ for naive reuse (Appendix~\ref{app:accum}).
These results complement the hidden-state drift analysis above by quantifying deviation from Full Rebuild and its output-level effects across memory-content update patterns.

\begin{table}[h]
\centering
\caption{
Analysis on Deviation from Full Rebuild under memory-content update patterns.
$\ourmodel$ keeps reconciled KV states close to full recomputation by recomputing affected KV ranges.}
\label{tab:ana:approx}
\begin{adjustbox}{width=.99\linewidth}
\begin{tabular}{lccccc}
\toprule
\textbf{Update pattern} & \textbf{K cos} & \textbf{Hidden cos} & \textbf{KL (nats)} & \textbf{Top-1 agreement} & \textbf{Top-5 overlap} \\
\midrule
Reordering & $0.973 \pm 0.003$ & $0.985 \pm 0.001$ & $0.036 \pm 0.008$ & $0.80$ & $0.96$ \\
Insertion before reused ranges & $0.987 \pm 0.001$ & $0.990 \pm 0.001$ & $0.061 \pm 0.013$ & $1.00$ & $0.92$ \\
Eviction before reused ranges & $0.986 \pm 0.001$ & $0.984 \pm 0.001$ & $0.078 \pm 0.017$ & $0.80$ & $0.76$ \\
Content change & $0.997 \pm 0.001$ & $0.998 \pm 0.000$ & $0.006 \pm 0.003$ & $0.80$ & $1.00$ \\
Adapter switch & $0.999 \pm 0.002$ & $0.992 \pm 0.008$ & $0.009 \pm 0.010$ & $1.00$ & $0.88$ \\
\bottomrule
\end{tabular}
\end{adjustbox}
\end{table}

\paragraph{Memory scaling under increasing chunk counts.}
\label{app:analysis:scaling}
We further evaluate the scalability of $\ourmodel$ as the number of memory chunks increases under the ALFWorld memory composition.
Table~\ref{tab:app:chunk_scaling} reports the relative step cost and GPU memory usage with $10\%$ of memory chunks updated at each decision step.
As the number of chunks increases from $8$ to $128$, the step cost of $\ourmodel$ relative to Full Rebuild decreases from $0.330$ to $0.206$. 
At $128$ chunks, $\ourmodel$ requires $22.0$ GB of peak GPU memory, compared with $46.8$ GB for Full Rebuild.

We additionally measure the overhead of reconciliation planning and cache assembly as the memory context grows. 
Reconciliation planning increases from $37$ to $310$ microseconds, while cache assembly increases from $0.7$ to $5.7$ ms across the evaluated memory sizes.
These results indicate that $\ourmodel$ reduces redundant recomputation as the memory context grows, while maintaining limited overhead for planning and assembling reusable KV ranges.

\begin{table}[h]
\centering
\caption{
Analysis on memory scaling under increasing chunk counts.
$\ourmodel$ reduces step cost relative to Full Rebuild and maintains lower peak GPU memory usage by recomputing only affected KV ranges under a fixed memory-content update ratio.}
\label{tab:app:chunk_scaling}
\begin{adjustbox}{width=.60\linewidth}
\begin{tabular}{cccc}
\toprule
\textbf{Chunks ($N$)}
& \textbf{Relative cost}
& \multicolumn{2}{c}{\textbf{GPU peak (GB)}} \\
\cmidrule(rl){3-4}
& & \textbf{$\ourmodel$} & \textbf{Full Rebuild} \\
\midrule
8   & 0.330 & 7.1  & 7.6 \\
32  & 0.255 & 8.7  & 11.3 \\
128 & 0.206 & 22.0 & 46.8 \\
\bottomrule
\end{tabular}
\end{adjustbox}
\end{table}

\paragraph{Sensitivity to dependency density.}
We further analyze how dependencies among memory chunks affect KV reuse and recomputation cost.
Table~\ref{tab:app:dependency_density} reports the step cost as dependency density increases from $0\%$ to $100\%$.
Under schema-guided reconciliation, invalidation propagates from updated memory chunks along their declared dependencies.
Therefore, denser dependencies expand the affected KV ranges requiring local recomputation.

As dependency density increases from $0\%$ to $100\%$, the KV reuse ratio gradually decreases from $85.4\%$ to $61.3\%$, while the step cost relative to Full Rebuild increases from $0.247$ to $0.368$.
The intermediate densities of $25\%$ and $50\%$ exhibit reuse ratios of $81.8\%$ and $73.1\%$, respectively, indicating a gradual reduction in KV reuse as dependencies become denser.

\begin{table}[h]
\centering
\caption{
Analysis on KV reuse and step cost under increasing dependency density.
Denser dependencies expand the affected KV ranges requiring local recomputation, while $\ourmodel$ maintains lower step cost than Full Rebuild across the evaluated densities.}
\label{tab:app:dependency_density}

\small
\setlength{\tabcolsep}{5pt}
\renewcommand{\arraystretch}{1.1}

\begin{adjustbox}{max width=.95\linewidth}
\begin{tabular}{cccc}
\toprule
\textbf{Density (\%)}
& \textbf{Reuse ratio (\%)}
& \textbf{Relative cost}
& \textbf{Step cost (ms)} \\
\midrule
0   & $85.4 \pm 0.3$ & $0.247 \pm 0.016$ & 18 \\
25  & $81.8 \pm 1.5$ & $0.274 \pm 0.017$ & 21 \\
50  & $73.1 \pm 2.2$ & $0.304 \pm 0.014$ & 23 \\
100 & $61.3 \pm 7.5$ & $0.368 \pm 0.062$ & 27 \\
\bottomrule
\end{tabular}
\end{adjustbox}
\end{table}

\paragraph{Adapter pool scaling and switching overhead.}
We further evaluate the memory footprint and switching overhead of procedural adapters.
The adapter memory footprint increases approximately linearly with pool size, requiring about $9.18$ MB per adapter and $587.2$ MB for a pool of $64$ adapters.
The larger adapter pools are constructed by replicating the shared, bathrooms, bedrooms, and livingrooms adapters under distinct adapter identifiers.
Therefore, this experiment measures system-level memory and switching overhead rather than semantic interference among distinct procedural skills.

We additionally evaluate adapter-stable KV reconciliation under forced adapter switching at every decision step.
$\ourmodel$ maintains a KV reuse ratio of $77.9\%$, whereas disabling adapter-stable KV reconciliation
reduces reuse to zero and yields latency comparable to Full Rebuild.
This result shows that adapter-stable KV reconciliation preserves compatible KV states across frequent procedural adapter switches, avoiding unnecessary recomputation of the memory context.

\paragraph{Per-step latency breakdown.}
\label{app:analysis:latency}
We profile end-to-end step latency on ALFWorld, including memory retrieval and prompt construction, whereas benchmark inference latency is measured after the step input is finalized.
Table~\ref{tab:app:latency_breakdown} decomposes the measured step latency into memory retrieval and context construction, prompt construction, KV reconciliation, local recomputation, fresh-tail prefill, and decoding.
Both $\ourmodel$ and Full Rebuild use the same evaluation protocol, prompts, and decoding configuration.

The largest difference occurs in local recomputation, which decreases from $267.4$ ms for Full Rebuild to $39.5$ ms for $\ourmodel$.
Memory retrieval, prompt construction, KV reconciliation, and fresh-tail prefill show closely matched costs between the two methods.
The amortized decode cost is approximately $24.0$ ms per generated token for both methods.
Overall, the measured step latency decreases from $1{,}287.5$ to $1{,}043.3$ ms. 
In particular, the $227.9$ ms reduction in local recomputation accounts for nearly all of the $228.2$ ms reduction in TTFT.
This shows that the efficiency gain primarily comes from avoiding redundant memory-context recomputation rather than changes to other inference components.

This instrumented profiling experiment is intended to attribute latency to individual inference components, rather than reproduce the benchmark-level TTFT averages.
Therefore, its absolute latency values are not directly comparable with those reported in the main evaluation.

\begin{table}[h]
\centering
\caption{
Analysis on per-step latency across inference components.
$\ourmodel$ reduces local memory-context recomputation while maintaining comparable retrieval, prompt construction, fresh-tail prefill, and decoding costs to Full Rebuild.
Amortized decode cost is the decoding time after the first token divided by the total number of generated tokens.}
\label{tab:app:latency_breakdown}
\begin{adjustbox}{width=.75\linewidth}
\begin{tabular}{lrr}
\toprule
\textbf{Component}
& \textbf{$\ourmodel$}
& \textbf{Full Rebuild} \\
\midrule
Memory retrieval and context build (ms) & 0.6 & 0.6 \\
Prompt construction (ms)                & 2.3 & 2.4 \\
KV reconciliation (ms)                  & 0.6 & 0.6 \\
Local recomputation (ms)                & $\mathbf{39.5}$ & 267.4 \\
Fresh-tail prefill (ms)                 & 131.4 & 131.7 \\
Decode after the first token (ms)       & 868.9 & 884.8 \\
\midrule
\textbf{Total latency (ms)}
& $\mathbf{1{,}043.3}$ & 1,287.5 \\
\midrule
Generated tokens per step & 36.2 & 36.9 \\
Amortized decode cost (ms/token)    & 24.0 & 24.0 \\
\bottomrule
\end{tabular}
\end{adjustbox}
\end{table}

\subsection{Ablation}
Table~\ref{app:tab:ablation} provides a more detailed ablation of the main components of $\ourmodel$.
The full framework achieves the best result across all metrics, with $71.43\%$ SR, $0.18$ seconds per step, $75.30\%$ reuse, and $21.59$ ms TTFT.
This shows that the full framework improves both task-level robustness and serving-level efficiency.

Removing the memory schema mainly weakens the link between memory contents and KV ranges.
SR drops from $71.43\%$ to $62.67\%$, and reuse decreases from $75.30\%$ to $51.39\%$.
TTFT also increases from $21.59$ ms to $30.39$ ms.
This indicates that typed chunk anchoring is necessary for deciding which memory contents remain reusable after memory-content updates.
Without the schema, the system cannot reliably identify reusable chunks or map updated contents to affected KV ranges, leading to lower reuse and less effective action behavior.
The decrease in active action ratio and action diversity ratio further suggests that the agent uses a narrower and less productive set of actions when memory-to-cache alignment becomes unreliable.

Removing chunk-level KV reconciliation produces the largest degradation.
SR decreases to $59.40\%$, reuse drops to $42.75\%$, latency increases to $0.47$ seconds per step, and TTFT rises to $45.10$ ms.
This variant still has memory chunks, but it no longer localizes insertions, evictions, and reorderings to affected KV ranges.
As a result, the system must rely on broader recomputation or less stable reuse when the memory context changes.
The large TTFT increase shows that chunk-level reconciliation is the main source of prefill-cost reduction, while the SR drop shows that localized cache updates also matter for decision quality.

Removing adapter-stable KV reconciliation yields a different failure mode.
Reuse remains relatively high at $66.40\%$, but SR still drops to $64.85\%$ and TTFT increases to $39.75$ ms.
This shows that a high reuse ratio alone is not sufficient.
When procedural memory changes the active adapter, KV states computed under the previous adapter may no longer be compatible with the current procedural configuration.
Thus, adapter-stable reconciliation improves the compatibility of reused KV states with the active procedural memory.
The gap between this variant and the full framework supports the role of adapter-aware compatibility checks in maintaining stable action decisions.

Overall, the ablation results separate the roles of the three components.
The memory schema identifies which contents can be reused, chunk-level KV reconciliation localizes memory-content updates to affected ranges, and adapter-stable KV reconciliation prevents incompatible reuse under procedural changes.
Together, these components allow $\ourmodel$ to reduce inference cost while preserving task success and action-level robustness.

\begin{table*}[!h]
\centering
\caption{Detailed ablation on $\ourmodel$. `w/o.' denotes the removal of a component.}
\label{app:tab:ablation}
\resizebox{\textwidth}{!}{%
\begin{tabular}{lcccccc}
\toprule
\textbf{Method} 
& \textbf{SR (\%)} 
& \textbf{Latency (s/step)}
& \textbf{Reuse ratio(\%)}
& \textbf{Active ratio(\%)}
& \textbf{Diversity ratio(\%)}
& \textbf{TTFT (ms)} \\
\midrule
\textbf{\ourmodel} 
& $\mathbf{71.43 \pm 2.05}$ 
& $\mathbf{0.18 \pm 0.02}$ 
& $\mathbf{75.30 \pm 4.33}$ 
& $\mathbf{67.69 \pm 2.27}$ 
& $\mathbf{71.43 \pm 3.32}$ 
& $\mathbf{21.59 \pm 2.53}$ \\
w/o Memory Schema 
& $62.67 \pm 3.49$ 
& $0.38 \pm 0.12$ 
& $51.39 \pm 5.84$ 
& $57.20 \pm 4.12$ 
& $61.00 \pm 5.03$ 
& $30.39 \pm 3.24$ \\
w/o Chunk-Level KV Reconciliation 
& $59.40 \pm 4.27$ 
& $0.47 \pm 0.05$ 
& $42.75 \pm 6.20$ 
& $55.10 \pm 4.75$ 
& $59.25 \pm 5.61$ 
& $45.10 \pm 1.21$ \\
w/o Adapter-Stable KV Reconciliation 
& $64.85 \pm 3.58$ 
& $0.28 \pm 0.03$ 
& $66.40 \pm 4.92$ 
& $62.71 \pm 3.66$ 
& $65.15 \pm 4.44$ 
& $39.75 \pm 2.22$ \\
\bottomrule
\end{tabular}%
}
\end{table*}

\clearpage
\newpage
\section*{NeurIPS Paper Checklist}

\begin{enumerate}

\item {\bf Claims}
    \item[] Question: Do the main claims made in the abstract and introduction accurately reflect the paper's contributions and scope?
    \item[] Answer: \answerYes{} %
    \item[] Justification: The claims in the abstract and introduction are aligned with the proposed method and the reported experimental results.

\item {\bf Limitations}
    \item[] Question: Does the paper discuss the limitations of the work performed by the authors?
    \item[] Answer: \answerYes{} %
    \item[] Justification: The paper discusses limitations and future directions of the proposed approach.

\item {\bf Theory assumptions and proofs}
    \item[] Question: For each theoretical result, does the paper provide the full set of assumptions and a complete (and correct) proof?
    \item[] Answer: \answerNA{} %
    \item[] Justification: The paper does not present formal theoretical results or proofs.

    \item {\bf Experimental result reproducibility}
    \item[] Question: Does the paper fully disclose all the information needed to reproduce the main experimental results of the paper to the extent that it affects the main claims and/or conclusions of the paper (regardless of whether the code and data are provided or not)?
    \item[] Answer: \answerYes{} %
    \item[] Justification: The paper provides implementation details, experimental settings, and supplementary materials to support reproducibility.

\item {\bf Open access to data and code}
    \item[] Question: Does the paper provide open access to the data and code, with sufficient instructions to faithfully reproduce the main experimental results, as described in supplemental material?
    \item[] Answer: \answerYes{} %
    \item[] Justification: The source code and reproduction instructions are provided in the supplementary materials.

\item {\bf Experimental setting/details}
    \item[] Question: Does the paper specify all the training and test details (e.g., data splits, hyperparameters, how they were chosen, type of optimizer) necessary to understand the results?
    \item[] Answer: \answerYes{} %
    \item[] Justification: The paper specifies the main experimental settings, datasets, models, hyperparameters, and evaluation protocols.

\item {\bf Experiment statistical significance}
    \item[] Question: Does the paper report error bars suitably and correctly defined or other appropriate information about the statistical significance of the experiments?
    \item[] Answer: \answerYes{} %
    \item[] Justification: The paper reports mean results with standard deviations over multiple runs where applicable.

\item {\bf Experiments compute resources}
    \item[] Question: For each experiment, does the paper provide sufficient information on the computer resources (type of compute workers, memory, time of execution) needed to reproduce the experiments?
    \item[] Answer: \answerYes{} %
    \item[] Justification: The paper reports the main software and hardware configuration used in the experiments.
    
\item {\bf Code of ethics}
    \item[] Question: Does the research conducted in the paper conform, in every respect, with the NeurIPS Code of Ethics \url{https://neurips.cc/public/EthicsGuidelines}?
    \item[] Answer: \answerYes{} %
    \item[] Justification: The research conforms to the NeurIPS Code of Ethics.

\item {\bf Broader impacts}
    \item[] Question: Does the paper discuss both potential positive societal impacts and negative societal impacts of the work performed?
    \item[] Answer: \answerNA{} %
    \item[] Justification: This work is primarily methodological and is evaluated in simulated benchmarks and controlled robot environments. We do not anticipate direct societal deployment or direct negative societal impacts from the work as presented. Any real-world deployment of embodied agents based on this approach would require appropriate safety validation and human oversight.
    
\item {\bf Safeguards}
    \item[] Question: Does the paper describe safeguards that have been put in place for responsible release of data or models that have a high risk for misuse (e.g., pre-trained language models, image generators, or scraped datasets)?
    \item[] Answer: \answerNA{} %
    \item[] Justification: The paper does not release high-risk models, datasets, or assets requiring special safeguards.

\item {\bf Licenses for existing assets}
    \item[] Question: Are the creators or original owners of assets (e.g., code, data, models), used in the paper, properly credited and are the license and terms of use explicitly mentioned and properly respected?
    \item[] Answer: \answerYes{} %
    \item[] Justification: The paper cites the existing datasets, models, benchmarks, software packages and their licenses used in the experiments.

\item {\bf New assets}
    \item[] Question: Are new assets introduced in the paper well documented and is the documentation provided alongside the assets?
    \item[] Answer: \answerNA{} %
    \item[] Justification: The paper does not release new datasets or models, but the supplementary materials document the code and experimental scripts.

\item {\bf Crowdsourcing and research with human subjects}
    \item[] Question: For crowdsourcing experiments and research with human subjects, does the paper include the full text of instructions given to participants and screenshots, if applicable, as well as details about compensation (if any)? 
    \item[] Answer: \answerNA{} %
    \item[] Justification: The paper does not involve crowdsourcing or research with human subjects.

\item {\bf Institutional review board (IRB) approvals or equivalent for research with human subjects}
    \item[] Question: Does the paper describe potential risks incurred by study participants, whether such risks were disclosed to the subjects, and whether Institutional Review Board (IRB) approvals (or an equivalent approval/review based on the requirements of your country or institution) were obtained?
    \item[] Answer: \answerNA{} %
    \item[] Justification: The paper does not involve human subjects, so IRB approval is not applicable.

\item {\bf Declaration of LLM usage}
    \item[] Question: Does the paper describe the usage of LLMs if it is an important, original, or non-standard component of the core methods in this research? Note that if the LLM is used only for writing, editing, or formatting purposes and does \emph{not} impact the core methodology, scientific rigor, or originality of the research, declaration is not required.
    \item[] Answer: \answerYes{} %
    \item[] Justification: The paper uses foundation models as embodied-agent backbones and describes the corresponding model, decoding, prompt, and adapter settings in the appendix.

\end{enumerate}

\end{document}